\documentclass[times,article,twocolumn,final]{ieeeaccess}
\usepackage{cite}
\usepackage{amsmath,amssymb,amsfonts}
\usepackage{algorithmic}
\usepackage{graphicx}
\usepackage{textcomp}

\usepackage{framed,multirow}
\usepackage{multirow}
\usepackage[table,xcdraw]{xcolor}
\usepackage{tablefootnote}
\usepackage{colortbl}
\usepackage{amssymb}
\usepackage{amsmath} 
\usepackage{latexsym}
\usepackage{booktabs}

\usepackage{url}
\usepackage{xcolor}
\definecolor{newcolor}{rgb}{.8,.349,.1}

\usepackage{color,soul}

\usepackage{stfloats}
\usepackage{float}
\usepackage{comment}

\usepackage{acronym}
\usepackage{enumitem}
\usepackage{graphicx}

\newacro{XASR}[XASR+]{Adaptive Sparse Representation}
\newacro{AI}[AI]{Artificial Intelligence}
\newacro{AM}[AM]{additive manufacturing}
\newacro{ANN}[ANN]{artificial neural network}
\newacro{AP}[AP]{Average Precision}
\newacro{AUC}[AUC]{Area Under Curve}
\newacro{BOW}[BoW]{Bag of Words}
\newacro{CAE}[CAE]{convolutional autoencoder}
\newacro{CNN}[CNN]{convolutional neural network}
\newacro{CST}[CST]{Cascaded Structure Tensor}
\newacro{CV}[CV]{computer vision}
\newacro{DL}[DL]{deep learning}
\newacro{DOG}[DoG]{Difference of Gaussians}
\newacro{FCM}[FCM]{fuzzy C-means clustering}
\newacro{FCNN}[FCNN]{fully-connected neural network}
\newacro{FNN}[FNN]{feed-forward neural network}
\newacro{FPN}[FPN]{Feature Pyramid Network}
\newacro{FPR}[FPR]{false positive rate}
\newacro{GAN}[GAN]{Generative Adversarial Network}
\newacro{GMM}[GMM]{Gaussian Mixture Model}
\newacro{HOG}[HOG]{Histogram of Oriented Gradients}
\newacro{IOU}[IoU]{Intersection over Union}
\newacro{KNN}[k-NN]{k-nearest neighbors}
\newacro{LBP}[LBP]{Local Binary Pattern}
\newacro{MCC}[MCC]{Mathews Correlation Coefficient}
\newacro{MCT}[$\mu$-CT]{micro computed tomography}
\newacro{MAP}[mAP]{Mean Average Precision}
\newacro{MGRTS}[MGRTS]{Mask Gradient Response-based Threshold Segmentation}
\newacro{ML}[ML]{machine learning}
\newacro{MLP}[MLP]{multi-layer perceptron}
\newacro{MOPE}[MOPE]{Multi-Output Proximity Embedding}
\newacro{MSE}[MSE]{mean squared error}
\newacro{NB}[NB]{Naive Bayes}
\newacro{NN}[NN]{neural network}
\newacro{PANET}[PANet]{Path Aggregation Network} 
\newacro{PCA}[PCA]{principal component analysis}
\newacro{RELU}[ReLU]{radial basis function}
\newacro{RBF}[RBF]{radial basis function}
\newacro{RF}[RF]{random forest}
\newacro{ROC}[ROC]{Receiver Operator Characteristics}
\newacro{ROI}[RoI]{Region of Interest}
\acrodefplural{ROI}[RoIs]{Regions of Interest}
\newacro{SAE}[SAE]{sparse auto-encoder}
\newacro{SIFT}[SIFT]{Scale Invariant Feature Transform}
\newacro{SLIC}[SLIC]{simple linear iterative clustering}
\newacro{SRC}[SRC]{sparse representation-based classification}
\newacro{SURF}[SURF]{Speeded Up Robust Features}
\newacro{SVM}[SVM]{Support Vector Machine}
\newacro{TNR}[TNR]{true negative rate}
\newacro{TPR}[TPR]{true positive rate}
\newacro{XCT}[XCT]{X-ray computed tomography}
\newacro{WGAN}[WGAN]{Wasserstein Generative Adversarial Network}

\def\BibTeX{{\rm B\kern-.05em{\sc i\kern-.025em b}\kern-.08em
    T\kern-.1667em\lower.7ex\hbox{E}\kern-.125emX}}
\begin{document}
\history{}
\doi{}

\title{\centering Computer Vision on X-ray Data in Industrial Production  and Security Applications: A Comprehensive Survey}
\author{\centering \uppercase{Mehdi Rafiei}\authorrefmark{1},
\uppercase{Jenni Raitoharju}\authorrefmark{2}  and \uppercase{Alexandros Iosifidis}\authorrefmark{3}
}
\vspace{5mm}

\address[1]{\centering Aarhus University, Nordre Ringgade 1, Aarhus C 8000, Denmark (e-mail: rafiei@ece.au.dk)}
\address[2]{\centering University of Jyväskylä, Mattilanniemi 2, 40100 Jyväskylä, Finland (e-mail: jenni.k.raitoharju@jyu.fi)}
\address[3]{\centering Aarhus University, Nordre Ringgade 1, Aarhus C 8000, Denmark (e-mail: ai@ece.au.dk)}


\vspace{5mm}

\begin{abstract}
X-ray imaging technology has been used for decades in clinical tasks to reveal the internal condition of different organs, and in recent years, it has become more common in other areas such as industry, security, and geography. The recent development of computer vision and machine learning techniques has also made it easier to automatically process X-ray images and several machine learning-based object (anomaly) detection, classification, and segmentation methods have been recently employed in X-ray image analysis. Due to the high potential of deep learning in related image processing applications, it has been used in most of the studies. This survey reviews the recent research on using computer vision and machine learning for X-ray analysis in industrial production and security applications and covers the applications, techniques, evaluation metrics, datasets, and performance comparison of those techniques on publicly available datasets. We also highlight some drawbacks in the published research and give recommendations for future research in computer vision-based X-ray analysis.
\end{abstract}

\begin{keywords}
Computer vision, Deep learning, X-ray, Industrial applications, Security applications
\end{keywords}

\titlepgskip=-15pt

\maketitle

\section{Introduction}

The need of having a non-destructive procedure for examining the interior of objects to assess their structural patterns or constituent contents has resulted in many applications of X-ray technology in different fields. While the medical field was one of the first to use the technology for assessing the inner parts of the body \cite{medical}, the use of X-ray technology is expanding considerably for industrial and security purposes \cite{C2, S4}. Factories can now assess whether there are anomalies or defects inside a product without destroying it \cite{C8}, and border patrol officers at security gates can check for forbidden objects inside baggages without opening them \cite{S3}. 

\begin{table*}[]
\caption{\label{Surveys} \vtop{\hbox{\strut Overview of available surveys on computer vision for industrial and security X-ray applications}}}
\centering
\footnotesize

\begin{tabular}{lllllllllllllllll}
\hline
\cellcolor[HTML]{A6A6A6} & \cellcolor[HTML]{A6A6A6} & \multicolumn{7}{c}{\cellcolor[HTML]{A6A6A6}Covered X-ray applications} & \multicolumn{7}{c}{\cellcolor[HTML]{A6A6A6}Covered public datasets} \\
\multirow{-2}{*}{\cellcolor[HTML]{A6A6A6}Survey}  & \multirow{0}{*}{}{\cellcolor[HTML]{A6A6A6}\rotatebox[origin=c]{270}{Publication year}} & \cellcolor[HTML]{A6A6A6}\rotatebox[origin=c]{270}{Additive Manufacturing} & \cellcolor[HTML]{A6A6A6}\rotatebox[origin=c]{270}{Casting} & \cellcolor[HTML]{A6A6A6}\rotatebox[origin=c]{270}{Welding} & \cellcolor[HTML]{A6A6A6}\rotatebox[origin=c]{270}{Security} & \cellcolor[HTML]{A6A6A6}\rotatebox[origin=c]{270}{Electronic industry} & \cellcolor[HTML]{A6A6A6}\rotatebox[origin=c]{270}{Material sciences} & \cellcolor[HTML]{A6A6A6}\rotatebox[origin=c]{270}{Others} & \cellcolor[HTML]{A6A6A6}\rotatebox[origin=c]{270}{CoCr AM XCT \cite{D1-CoCr-Introduce}} & \cellcolor[HTML]{A6A6A6}\rotatebox[origin=c]{270}{GDXray \cite{gdxray}} & \cellcolor[HTML]{A6A6A6}\rotatebox[origin=c]{270}{SIXray \cite{sixray}} & \cellcolor[HTML]{A6A6A6}\rotatebox[origin=c]{270}{OPIXray \cite{opixray}} & \cellcolor[HTML]{A6A6A6}\rotatebox[origin=c]{270}{PIDray \cite{PIDray}} & \cellcolor[HTML]{A6A6A6}\rotatebox[origin=c]{270}{HiXray \cite{HiXray}} & \cellcolor[HTML]{A6A6A6}\rotatebox[origin=c]{270}{CLCXray \cite{CLCXray}} \\ \hline
Mery \cite{mery2015computer} & 2015 & & \checkmark & \checkmark & \checkmark & & & & & \checkmark & &  & &  & \\
\rowcolor[HTML]{D9D9D9} 
Hou et al. \cite{survey_welding} & 2020 & & & \checkmark & & & & & & & & & &  & \\
Mery at al. \cite{survey_security_CV} & 2020 & & & & \checkmark & & & & & \checkmark & \checkmark & \multicolumn{2}{l}{}  &  &\\
\rowcolor[HTML]{D9D9D9} 
Akcay and Breckon \cite{survey_security_DL_2022} & 2022 & & & & \checkmark & & & & & \checkmark & \checkmark & \checkmark &  &  &\\
Our survey & 2022 & \checkmark & \checkmark & \checkmark & \checkmark & \checkmark & \checkmark & \checkmark & \checkmark & \checkmark & \checkmark & \checkmark & \checkmark & \checkmark & \checkmark \\ \hline  
\end{tabular}
\end{table*}
\begin{figure*}[b]
\centering
\includegraphics[width=\textwidth]{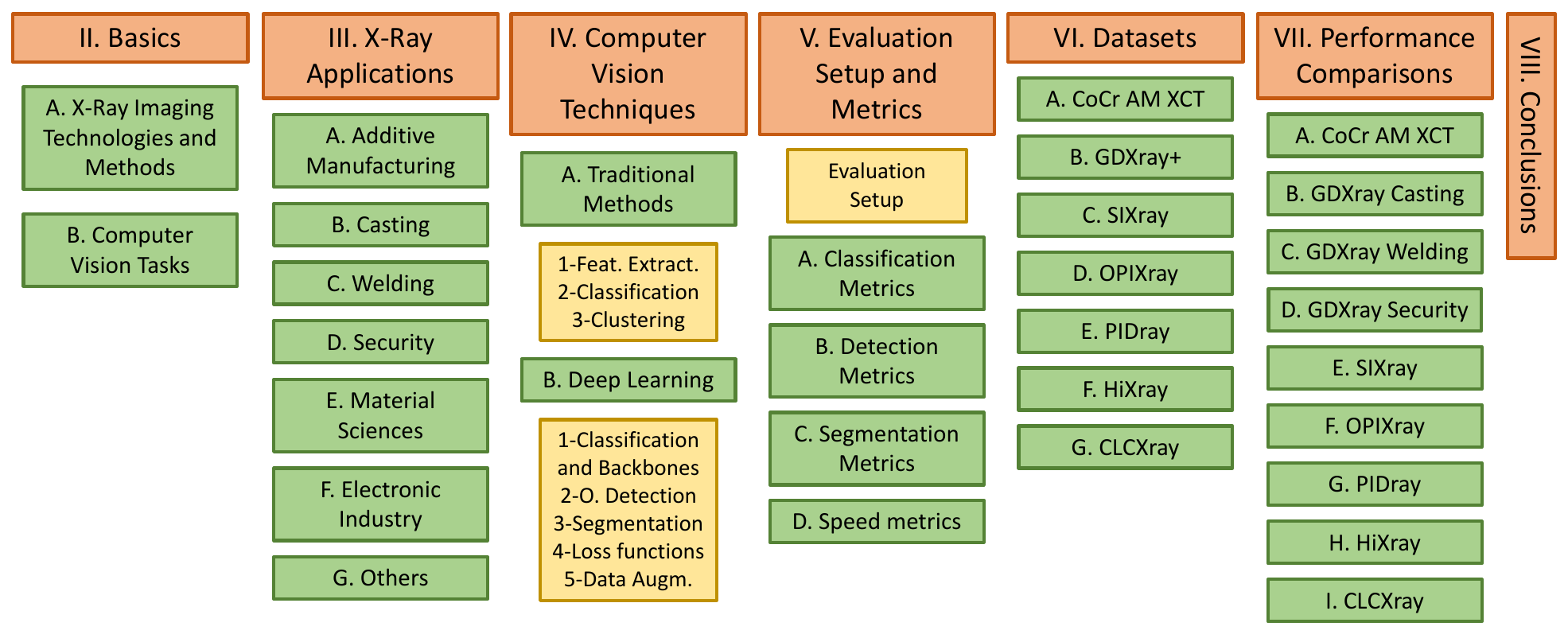}
\caption{Paper outline.}
\label{fig:outline}
\end{figure*}

Considering the need for fast production lines in competitive industries and issues related to human-based image assessment, such as subjectivity and tiredness, the necessity of automatic and reliable image processing methods is obvious. Efficient automatic techniques for processing X-ray data are needed also at security gates for baggage checking due to the increasing number of travelers. Recent advances in \ac{CV}, \ac{ML}, and \ac{DL} have the potential to provide efficient and reliable solutions to X-ray-based automatic and real-time object, anomaly, or defect detection and recognition.

As the need for automatic X-ray applications is growing in industrial production and security fields, several works have proposed different \ac{CV}-based techniques for processing the data. To get an overall understanding of the current state of the research, an extensive review of the published articles is needed. There are also previously published related surveys: A book by Mery \cite{mery2015computer} discusses widely
computer vision algorithms for industrially relevant applications of X-ray testing, but does not cover the recent deep learning based-advances. A survey by Hou et al. \cite{survey_welding} focuses on computer-aided weld defect detection from radiography images. Surveys by Mery et al. \cite{survey_security_CV} and Akcay and Breckon \cite{survey_security_DL_2022} focus on computer vision for security applications. A summary of the available surveys on this topic is presented in Table \ref{Surveys}, where $\checkmark$'s indicate the covered topics. We note that the previously published surveys in these topics cover only a specific application and/or just a limited type of \ac{CV} methods. Furthermore, despite their obvious importance, the previously published surveys do not comprehensively discuss evaluation setups and metrics for different \ac{CV} tasks in X-ray analysis.

We provide an extensive review of recent \ac{CV} methods for X-ray image processing in industrial production and security applications. We aim to provide a full picture of both X-ray data and applications as well as \ac{CV} tasks and techniques to be useful for experts, as well as for readers with no previous experience on one or both of the sides. We also describe evaluation metrics and public datasets for the covered applications. In addition, we summarize the performance evaluation of many of the existing methods conducted on the public datasets for easier comparison of different techniques. We observed some common limitations, in particular in the experimental protocols applied on the reviewed works, which we bring up in our survey. Furthermore, we give recommendations for future research in computer vision-based X-ray analysis to remedy such problems.

The remainder of the paper is organized as follows. An introduction to X-ray imaging and \ac{CV} tasks is provided in Section~\ref{sec:basics}. In Section~\ref{sec:fields}, we review works in the field of \ac{CV} for X-ray data, categorized into seven different research fields (additive manufacturing, casting, welding, security, electronic industry, material sciences, and others). Section~\ref{sec:cvtechniques} introduces different \ac{CV} methods applied on X-ray data divided into traditional and \ac{DL}-based methods. Section~\ref{sec:evaluation} discusses how to evaluate the proposed methods for different applications. Section~\ref{sec:datasets} introduces the publicly-available datasets.  
The performance comparison of the existing methods on public datasets is provided in Section~\ref{sec:comparison}. Finally, the paper is concluded in Section~\ref{sec:conclusion} and recommendations for future research are given. The paper outline and context are shown in an organization chart in Fig.~\ref{fig:outline}. The important abbreviations used throughout the
paper are listed in Table~\ref{tab:abbreviations}.

\begin{table}
    \centering
    \caption{List of important abbreviations used throughout the paper in alphabetical order.}
    \label{tab:abbreviations}
    \begin{tabular}{@{}ll@{}}
        \toprule
        \textbf{Abbreviation} & \textbf{Definition} \\
        \midrule
        \acs{AM} & \Acl{AM} \\
        \acs{AP} & \Acl{AP} \\
        \acs{AUC} & \Acl{AUC} \\
        \acs{BOW} & \Acl{BOW} \\
        \acs{CNN} & \Acl{CNN} \\
        \acs{CST} & \Acl{CST} \\
        \acs{CV} & \Acl{CV} \\
        \acs{DL} & \Acl{DL} \\
        \acs{GAN} & \Acl{GAN} \\
        \acs{FNN} & \Acl{FNN} \\
        \acs{FPN} & \Acl{FPN} \\
        \acs{IOU} & \Acl{IOU} \\
        \acs{MAP} & \Acl{MAP} \\
        \acs{ML} & \Acl{ML} \\
        \acs{NN} & \Acl{NN} \\
        PDN & Part-based Detection Network \\
        ResNet & Residual neural network \\
        R-CNN & Region-based CNN \\
        \acs{ROC} & \Acl{ROC} \\
        \acs{ROI} & \Acl{ROI} \\
        \acs{SIFT} & \Acl{SIFT} \\
        \acs{SRC} & \Acl{SRC} \\
        \acs{SURF} & \Acl{SURF} \\
        SSD & Single Shot Multibox Detector \\
        \acs{SVM} & \Acl{SVM} \\
        \acs{XASR} & \Acl{XASR} \\
        \acs{XCT} & \Acl{XCT} \\
        YOLO & You Only Look Once \\
        \acs{MCT} & \Acl{MCT} \\
        
        \bottomrule
    \end{tabular}
\end{table}

\section{Basics of X-ray Imaging and Computer Vision Tasks}
\label{sec:basics}

In order to provide a basic understanding of both X-ray imaging and computer vision tasks and to make it easier to follow the rest of the paper for readers unfamiliar with the topics, this section provides basic information about available techniques and related definitions.

\subsection{X-ray Imaging Technologies and Methods}

\begin{figure}[!t]

\includegraphics[width=\columnwidth]{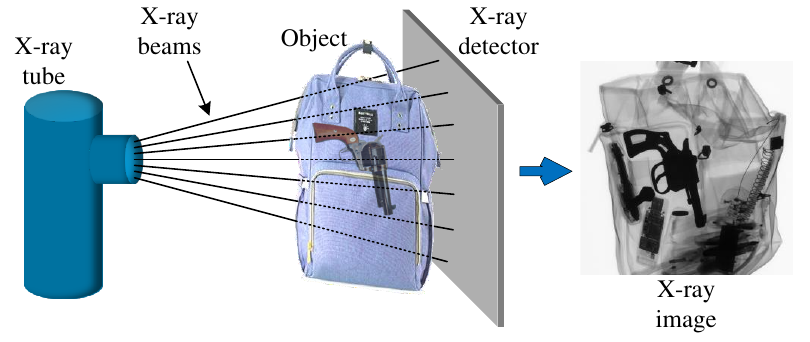}
\caption{X-ray imaging technology.}
\label{X-ray-imaging}
\end{figure}

While visual imaging sensors measure light reflected on surfaces to capture their color image, in X-ray imaging, the ionizing beams are generated by X-ray tubes and penetrated through the scanned object to be detected by the detectors on the other side of the object (Fig.~\ref{X-ray-imaging}). Depending on the mass density of the exposed object, the X-ray signal can be attenuated, which leads to a lower intensity on the detector. In other words, the measured intensity is inversely proportional to the material density. Therefore, X-ray imaging can be used to carry out non-destructive assessments, when mass density is a parameter of interest.

\begin{figure}[!t]

\includegraphics[width=\columnwidth]{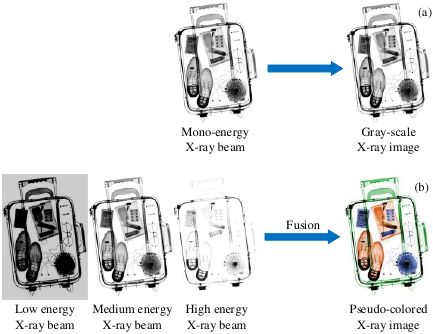}
\caption{Different X-ray imaging techniques based on the number of beam sources: a) mono-energy, b) multi-energy.}
\label{gray-color-img}
\end{figure}

X-ray imaging techniques can be categorized based on the number of energy levels and the view they use. In terms of energy level, X-ray imaging techniques can be divided into mono- or multi-energy levels (irrespective of the view or the number of X-ray beams). In mono-energy X-ray imaging technology, only one energy level is used for the radiated beams \cite{advanced-X-ray}, which provides gray-scale images (see Fig.~\ref{gray-color-img}-a) according to the object mass density. This is a suitable X-ray imaging technology when dealing with mono-material objects (e.g., in additive manufacturing). Dual- and multi-energy X-ray imaging technologies use several energy levels to provide several X-ray images leading to a better understanding of the objects' density and effective atomic number \cite{multi-energy}. By using of look-up table \cite{look-up-table}, the measured values can be transferred to a pseudo-colored image of the object (see Fig.~\ref{gray-color-img}-b), where various colors are assigned to different types of material. Thus, when different types of material are assessed (e.g., in baggage security checks), these X-ray imaging technologies can provide more information and make it easier to analyze the inner structure of the objects.

\begin{figure}[!t]
\centering
\includegraphics[width=\columnwidth]{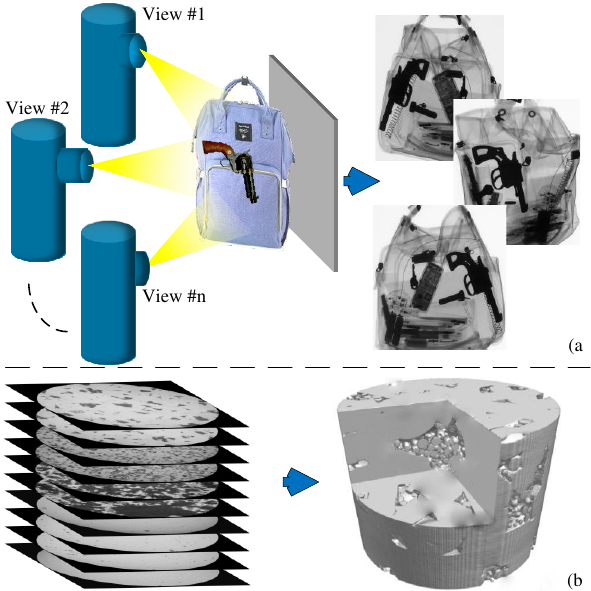}
\caption{Different X-ray imaging techniques based on the view: a) multi-view, b) 3D view.}
\label{view-img}
\end{figure}

In terms of the view, X-ray imaging techniques can be divided into 2D, multi-view, and 3D imaging categories. In 2D imaging, the X-ray beams are radiated by the X-ray tube to the object from only one direction producing 2D images. In multi-view imaging, the objects of interest are exposed to the X-ray beams from various angles \cite{multi-view} (see Fig.~\ref{view-img}-a), providing more information on the object facilitating the analyses. In 3D view imaging, the output is in a 3D form and it can be provided in different ways. One approach, also known as tomography, is to capture 2D X-ray images of different layers of the object and then stack them on top of each other (see Fig.~\ref{view-img}-b) to provide a 3D volume of it \cite{D1-CoCr-Introduce, D1-CoCr-Source, thompson2016amreview}. Another approach for 3D imaging is to combine multi-view imaging with image processing techniques to transfer the 2D images from different angles to a 3D volume \cite{E1}.

In the end, it should be mentioned that visual and X-ray imaging share some characteristics and challenges, such as noise, occlusion, and perspective. Therefore, many image processing methods developed for color images can be used also with X-ray images \cite{S23}.

\subsection{Computer Vision Tasks}

In this section, we briefly describe the main \ac{CV} tasks relevant to different X-ray applications. We start by defining the general terms: computer vision, machine learning, and deep learning. In many cases, the terms are interchangeable, but they still cover a different subset of methods. 

\begin{itemize}[leftmargin=*]
\item\emph{\Acf{CV}} is a sub-field of \ac{AI} that focuses on processing images and videos captured by a variety of sensors (e.g., visual cameras, X-ray imaging sensors, depth sensors). Many \ac{CV} methods are learning-based and thus also \ac{ML} methods, but there exist also non-learning-based \ac{CV} algorithms. 

\item\emph{\Acf{ML}} is a sub-field of \ac{AI}, where methods learn to perform a task without being explicitly programmed to do so. To be able to learn, most ML models need to be trained by using input-target output pairs. \Ac{ML} methods can be applied to many \ac{CV} tasks involving images/videos captured by a variety of sensors, but also to tasks involving various other data types, which are out of the scope of this paper.

\item\emph{\Acf{DL}} is a sub-field of \ac{ML}. \ac{DL} methods involve multiple layers of data transformations usually taking the form of neural layers to progressively extract higher level and more complex patterns from data.
\end{itemize}

\begin{figure}[!t]
\centering
\includegraphics[width=\columnwidth]{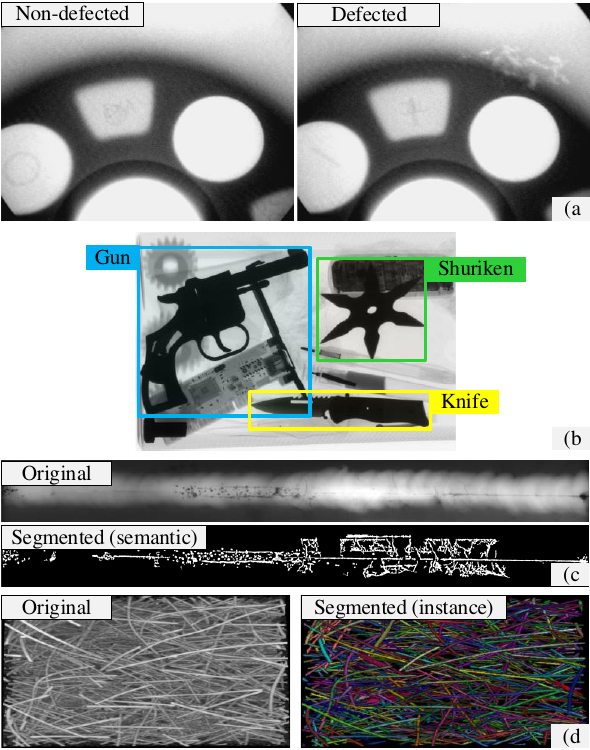}
\caption{Computer Vision task examples, a) Binary classification for casting defect detection \cite{gdxray-source}, b) Object detection in X-ray baggage security inspection \cite{gdxray-source}, c) Semantic segmentation of an X-ray welding image \cite{gdxray-source}, and d) Instance segmentation of glass fibers in industrial computed tomography \cite{M4}.}
\label{CV-tasks}
\end{figure}

\textbf{Classification} is a \ac{CV} task aiming at assigning a data sample (e.g., an image or a video) into one class included in a set of predefined classes. The classes can represent properties, such as intact/damaged, or types of depicted objects. In single-label classification (commonly simply referred to as classification), the classes are mutually exclusive and only one label is assigned to each sample, while in multi-label classification samples can be assigned a varying number of labels. Based on the number of classes, (single-label) classification tasks can be categorized into binary and multi-class types. Binary classification refers to tasks with only two classes, while multi-class classification refers to tasks with more than two classes. Fig.~\ref{CV-tasks}-a shows an example of a binary classification task on casting defect detection, where the image on the left is classified as non-defected and the image on the right as defected.

\textbf{Object detection} \cite{zhao2019object} aims at both localizing and identifying each object in an image/video. Bounding boxes are commonly used to represent where each object is in the image, while the identification can be seen as classification of the image patch depicting the object. Object detection allows counting different objects or following specific objects in videos. Fig.~\ref{CV-tasks}-b illustrates the output of object detection applied on an X-ray baggage security image. It can be seen that the trained \ac{ML} model detected three types of objects of interest and put bounding boxes around them. 

In \textbf{semantic segmentation} \cite{wang2018understanding}, the goal is to assign each pixel in an image to a class. Compared to classification and object detection tasks that provide the overall class or bounding boxes around the detected objects, semantic segmentation delivers an exact outline of the objects/content from different classes. Fig.~\ref{CV-tasks}-c presents an example of a semantically segmented X-ray welding image. In the segmented image, the exact pixel-wise locations of the defects are marked. \textbf{Instance segmentation} \cite{yang2019video} is a slightly different task, where the goal is to segment the image according to the different instances of the same class, e.g., to count the instances. In \cite{M4}, instance segmentation was applied for segmenting glass fibers in industrial computed tomography as illustrated in Fig.~\ref{CV-tasks}-d.  

While most \ac{CV}-based X-ray image analysis methods proposed in the literature can be categorized directly as methods targeting classification, object detection, or segmentation tasks, some works set their final objective beyond these tasks. For example, in \cite{O14}, object detection is used to identify incorrect assembly, missing assembly, or transposition of internal components of a product and, in \cite{A4}, defect segmentation is followed by estimation of different defect characteristics. In \cite{W13}, features extracted from X-ray images were used to predict geometrical parameters of welding as a regression task.

\section{Application Areas of X-ray Technology}
\label{sec:fields}
In this section, a short overview of different industrial production and security applications that use X-ray images along with computer vision techniques is provided. The main covered research topics are additive manufacturing, casting, welding, security, electronic industry, and material sciences. The related methods are categorized based on their underlying computer vision objective, i.e., classification, detection, or segmentation.

\subsection{Additive Manufacturing}
\Ac{AM} technology, which is also known as 3D printing \cite{A1}, is broadly used in diverse industrial applications with high material and geometric complexities, such as car manufacturing. \ac{AM} technology can use several techniques including directed energy deposition \cite{A4}, powder bed fusion \cite{A3}, binder jetting \cite{A4}, and additive friction stir deposition \cite{A4}, to built final or near-net-shape (i.e., initial roughly shaped) parts in a layer-by-layer manner directly from digital files. However, structural defects, such as pores, internal micro-cracks, air bubbles, surface pits, surface scratches, and porosity arrays, are inevitable in current AM processes \cite{A4}. Printing errors, cyberattacks, residual stress, powder materials, chamber environment, as well as printing parameters, namely heat source power, scan speed, hatch space, and layer thickness, are considered to be the possible reasons behind the mentioned defects \cite{A1}.

\Ac{XCT}, as a non-destructive evaluation process, is widely used in \ac{AM} processes to examine the internal and surface structure of produced parts to detect different defects \cite{thompson2016amreview}. Most of the \ac{CV}-based techniques used for \ac{AM} inspection aim at \textbf{segmenting} \ac{XCT} data in 2D or 3D \cite{D1-CoCr-Introduce, A1, A3}. In \cite{D1-CoCr-Introduce}, defect segmentation of cylindrical \ac{AM} specimens belonging CoCr AM XCT dataset (see Section \ref{ssec:amdataset}) was carried out using a local thresholding method on 2D slides. 3D defect segmentation of the same data based on 3D fully-convolutional network was carried out in \cite{A1}. In \cite{A3}, a 2D segmentation network was employed for automatic porosity segmentation of metallic \ac{AM} specimens. The \ac{XCT} data was processed as a stack of 2D images to provide porosity labels for the specimens, and different segmentation methodologies were evaluated. 

In \cite{A4}, an application going beyond the basic \ac{CV} tasks was proposed as an inspection pipeline and applied for defect characteristics and pore evolution analysis in a binder jetting copper \ac{AM} system.

\subsection{Casting}
Casting is a manufacturing process finding applications in complex industries, such as aerospace \cite{C2, C7} and automobile \cite{C3, C8} industries, and with materials, such as aluminum \cite{C1, C5} and titanium \cite{C2} alloys. Due to the limitations of the manufacturing techniques \cite{C1}, castings can host several defects, such as holes and flaws, gas cavities, shrinks, slags, cracks, high- and low-inclusions, wrinkles, casting fins, shrinkage-holes, and incomplete fusion \cite{C2, C3, C7}, which can lead to catastrophic failures of critical mechanical components \cite{C1, C8}. Therefore, it is crucial to implement a non-destructive testing system to detect internal and surface defects early in the manufacturing process to reduce the risks and save time and costs \cite{C3, C5}. 

To this end, X-ray imaging is becoming a useful technology to visualize the internal structure of castings and, combined with \ac{CV} methods, it allows for automatically assessing the products and detecting anomalies \cite{C6}. In casting assessment, most of the studies consider the problem as a binary task (e.g., binary classification or segmentation) to differentiate between defective and non-defective castings. There are also some cases where more than one type of defects define a multi-class detection or segmentation problem \cite{C2}.

Several works that frame the problem as a binary \textbf{classification} task \cite{C1,C06,C5,O3,wu2021ameliorated}, where the goal is to classify X-ray images into defective and non-defective classes, have been conducted in the automotive industry.
In aerospace industry applications, \cite{C7} evaluated and compared several traditional classification methods on casting defect classification of image patches in supporting plates in aeromotors.

In aerospace industry applications, most of the related research focuses on \textbf{detection} of casting defects in aeroengines. In \cite{C2}, defect candidate search and classification steps were applied to detect aerospace titanium casting defects. 
For detecting core failures in die casting, an unsupervised inspection framework was designed and introduced in \cite{C03}. A \ac{DL}-based detection method that tries to boost the detection performance at both data augmentation and algorithm levels was used in \cite{C6}. Casting defects were localized using several \ac{CNN} architectures that were trained on a relatively small dataset in \cite{C9}. As an alternative method, a classifier was trained on image series and a sliding window-based approach was applied for localization. To increase the safety in the construction of road-worthy metallic components, several casting defect detection \ac{CV} methods were assessed and compared in \cite{C3}.  The defect detection approach proposed in \cite{C8} trains the network simultaneously for detection and instance  segmentation on casting X-ray images. It was experimentally shown that simultaneous training for detection and instance segmentation led to a higher detection accuracy than training to detect alone. 

Semantic \textbf{segmentation} methods for casting inspection were assessed in \cite{C02, C04}. Authors in \cite{C02} used only realistically simulated X-ray data to train a network to perform semantic segmentation on cast aluminum parts. Large defect scale variation, small inter-class differences, and annotation uncertainty issues were tackled in \cite{C04} for defect semantic segmentation.

\subsection{Welding}
Welding, as a manufacturing process that joins materials by causing coalescence and melting two workpieces, plays a critical role in a variety of production applications, such as aircraft, shipbuilding, and automobile production \cite{W1}. However, the instability of welding process parameters, such as welding current, voltage, speed, and nozzle height, as well as the structural component deformation might cause defects in the joints \cite{W1, W2}. These defects can reduce the quality of the product by affecting strength, stiffness, safety, and durability and cause catastrophic damages \cite{W4, W8}. Due to the different welding energy sources, environments, chemical and physical processes, and joining materials, welding is considered a complex and not fully understood process \cite{W10}. Therefore, weld quality evaluation done by experts carries limitations, such as subjectivity, misinterpretation, and time consumption \cite{W14, W15}. To overcome the limitations and to achieve real-time geometrical parameter monitoring, automatic non-destructive techniques to assess welding quality are needed \cite{W13}. The combination of X-ray imaging technology and \ac{CV} methods is a commonly chosen approach in this regard. 

To \textbf{classify} different welding defects, a \ac{CNN}-based classification method was used in \cite{W01, W03}. In \cite{W03}, imbalanced class distribution was addressed by using resampling methods 
to create a balanced dataset. In \cite{W13}, an approach including \ac{FNN} and \ac{SVM} models 
was introduced for a laser welding process monitoring and defect recognition. 
In \cite{W5}, binary classifiers with low data requirement for generic automated surface inspection was presented. Defective weldings were classified in \cite{W10} employing a complex binary classifier consisting of an artificial \ac{NN} and a fuzzy logic system. A set of geometric features, such as shape measures (compactness, elongation, symmetry, etc.), was defined in \cite{W11} to characterize defects in X-ray data and then these features were used as inputs to a multi-class classifier that divides the problem into one versus one binary problems.

Studies in \cite{W4, W7, W2, W12} applied \textbf{detection} \ac{CV} tools on welding X-ray data. 
The class distribution of welding X-ray image sets was balanced in \cite{W4} using two data augmentation approaches, and then the balanced data and a feature extraction-based transfer learning method were used to train two deep models. The models were then combined to perform defect detection via dividing the image into sub-images and separately classifying each sub-image. The method in \cite{W7} uses a hybrid automatic detection scheme including a location extractor of weld region and a detector based on a binary classifier. The detector uses sliding window and the trained binary classifier to detect the defective parts. A \ac{DL}-based model was employed in \cite{W2} to automatically identify multiple welding defects and extract their location without any pre-processing. 
 
In \cite{W1}, high-precision automatic weld defect \textbf{segmentation} for small defects was achieved by employing a deep neural network and data augmentation. In \cite{W6}, an automatic welded joints' segmentation technique is introduced which can localize weld beads, segment discontinuities (as potential defects), and finally, extract the features to classify the discontinuities. In \cite{W14}, minimum intra-class and maximum inter-class variances were used to localize defects after applying a noise reduction method on X-ray data. Then, shape features were extracted and used to classify the defects. The potential weld joint defects were segmented using a background subtraction algorithm in \cite{W15}, and then defects' features including average gray-scale difference to the background, gray-scale standard deviation, and the defect area were extracted and used by a classifier to differentiate real defects from all potential ones. In \cite{W12}, defects were segmented by classifying each pixel using extracted feature vectors.

\subsection{Security}
\begin{figure}[!t]
\centering
\includegraphics[width=\columnwidth]{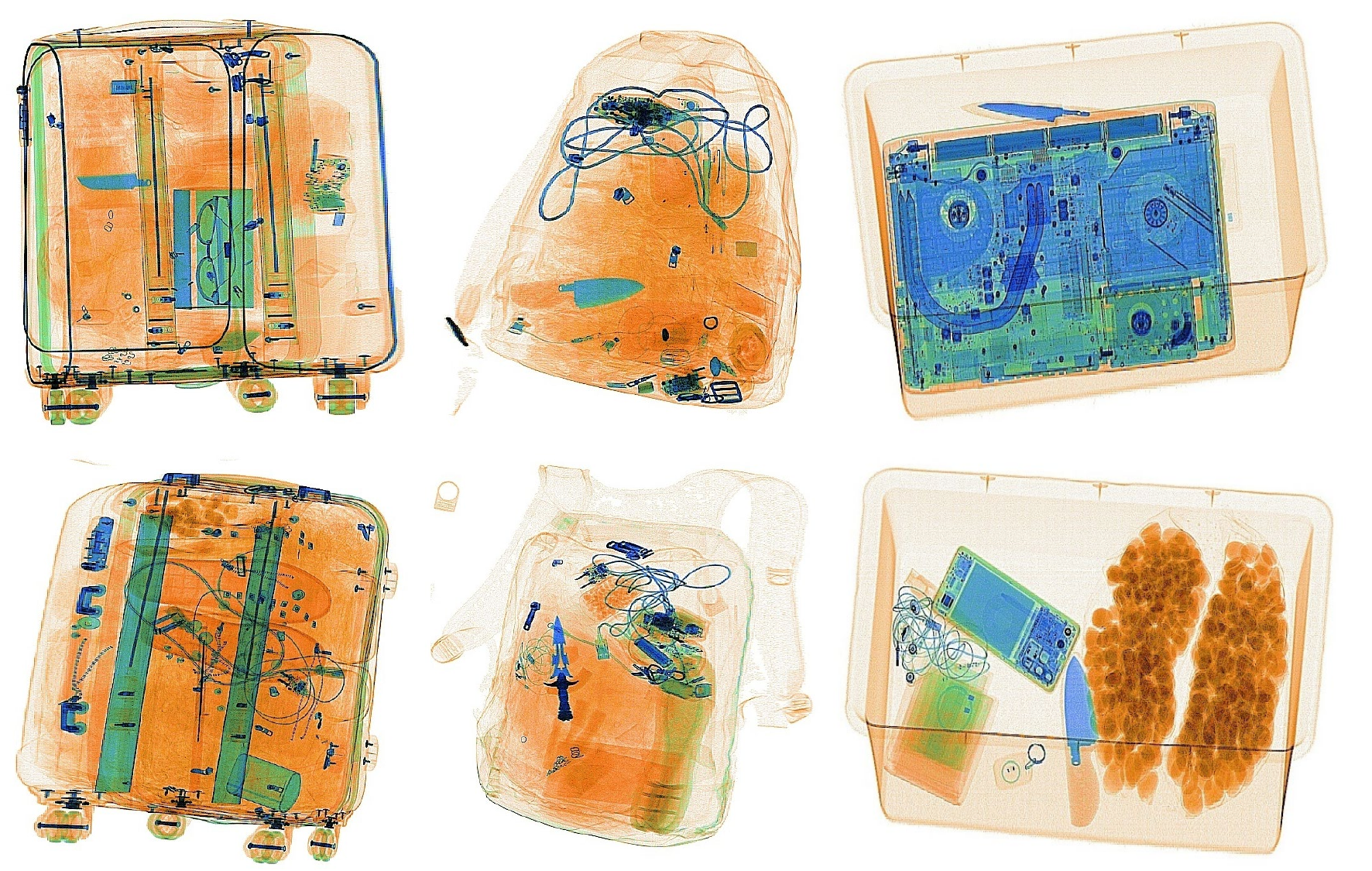}
\caption{Samples of X-ray security images, available in OPIXray data \cite{opixray}.}
\label{OPIX-img}
\end{figure}

One of the most common areas where X-ray imaging is used is baggage inspection at, for example, security gates on railway stations, subway stations, and airports, for detecting prohibited items and threats \cite{S1}. However, the inspection and threat detection in this context are usually performed by humans \cite{S1, S5}. Fatiguing work schedules, complexity in catching contraband items, inexperienced operators, and squeezed and overlapped items can be named as limitations of the human detection operation \cite{S3, S4, S6}. On the other hand, especially during busy hours, quick evaluation and detection are urgent to prevent any delay in the passengers' transportation schedules \cite{S9, S11}. All these together emphasize the necessity of reliable and time-efficient methods to do the detection automatically. Several datasets are available for developing \ac{CV}-based approaches as discussed in Section~\ref{sec:datasets}. Some sample images from one of the datasets, OPIXray \cite{opixray}, are shown in Fig. \ref{OPIX-img}.

To tackle the overlapping issue in X-ray images of tightly packed luggages, a multi-label \textbf{classification} network was used in \cite{S7} to recognize prohibited items such as guns, knives, scissors, etc. A multistage analyzer and classifier system was proposed in \cite{S16} to automatically perform threat recognition in different security monitoring environments to identify a wide range of firearm threats. 
To manage intra-class variability in contrast, pose, image size, and focal distance, a new representation approach was introduced in \cite{S24} to recognize objects. In \cite{S26}, a method was adopted for binary classification of firearms versus other objects in baggage security X-ray images. 
A deep \ac{NN} was employed in \cite{S25} as a security image classifier with the ability to overcome the data scarcity problem to classify among gun, knife, and electrical device classes. Ten diverse \ac{CV}-based strategies were investigated in \cite{S23} for object recognition in X-ray security imaging. In \cite{S15}, an anomaly detection method was proposed that categorizes anomalies in appearance and semantic anomalies. Unusual shape, texture, and density were considered as appearance anomalies and unfamiliar objects as semantic anomalies. 
A multi-scale \ac{CNN} architecture was used in \cite{S17} to discriminate materials into classes, such as metal or organic substances, using dual-energy X-ray scanner images.

Most studies in security applications aim at \textbf{detection} of threats. Some works focus on binary detection tasks to discriminate a specific class of objects, e.g., firearms/firearm components \cite{S12, S21, S22}.
The works in \cite{S18, S3, S6, S20, S22, S21} aim at recognizing multiple different threat categories, such as knives and guns, whereas the work in \cite{S10} aims at categorizing several object types, such as laptops and mobile phones, either as benign or anomalous.
An approach for extracting multi-level information and handling nonrigid deformations was employed in \cite{S014}. 
The problem of overlapping objects in X-ray security data was tackled in \cite{S015}. A patch-wise image classification method based on sparse representation of direction features was introduced in \cite{S8} and used for threatening object detection. The direction features were extracted to build a foreground dictionary used for assessing test data to detect the foreground. Another simple foreground-background segmentation technique based on color thresholds was applied in \cite{S9} as a preprocessing step for object detection in the X-ray images. 
To increase the trust in automatic detectors on baggage security imagery applications, a human-in-the-loop detection framework was presented in \cite{S14}. The framework gives a score to each prohibited item proposal and, based on the score, the baggage is assigned to safe, suspicious, or dangerous classes. In case of classification as suspicious, a human makes the final decision on the baggage.

Using synthetic data is a way to tackle the lack of data in security image detection \cite{S1, S5, S13}. In \cite{S1}, data for CNN training process was obtained using a method that generated X-ray security images with multiple prohibited items. The study in \cite{S5} used a data augmentation method that first generates several RGB images of prohibited items. Subsequently, the images are transformed into X-ray format and combined to different backgrounds. Synthetically composed X-ray images of transformed threats and backgrounds were used in \cite{S13} to overcome the high data requirement challenge in CNN training. To assess the approach, a CNN-based object recognition method was trained with both real and synthetic data and the results were compared, showing promising results for the combination of real and synthetic data.

Some studies on automatic X-ray security data analysis applied image \textbf{segmentation}. The work in \cite{S2} presented an automatic segmentation method for security screening that first enhances images to improve performance and then applies color-based pixel segmentation to distinguish diverse materials (organic, inorganic, mixed, and opaque objects) from the background. A method using a DL model as a robust feature extractor and an adversarial auto-encoder to classify images into organic and inorganic classes considering the overlap among the materials was proposed in \cite{S4}. To find the most suitable object level and sub-component level anomaly detection strategy, several segmentation methodologies were assessed in \cite{S11}. Their performances were analyzed by applying them to an extensive dataset focusing on electronic items.

\subsection{Material Sciences}
Following the recent developments in CV capabilities in combination with diverse X-ray imaging technologies, a new growing research topic focuses on the analysis of various materials using CV methods on X-ray images. It should be mentioned that the term material science here refers to the studies that focus on the properties of materials, not the production process of different materials. These methods can be employed to achieve realistic textile composite finite element models \cite{M2}, fiber extraction models \cite{M4}, links between microstructures and physical properties \cite{M5}, and characterization and mappings of materials \cite{M7}. 

Atomic resolution images of materials can be obtained by illuminating particles with random orientation with an X-ray free-electron laser beam and collecting of the scattering patterns.
In \cite{M1}, neural networks were used for binary \textbf{classification} of such diffraction patterns of non-crystalline objects into single hit or non-single hit classes. 

For \textbf{detection} of small-sized and dense void and inclusion defects in spacecraft composite structures, transfer learning and domain adaptation were used in \cite{M03}. In \cite{M04}, detection of internal defects of aluminum conductor composite core was performed as a patch-wise classification task.

Several works apply 2D or 3D \textbf{segmentation} techniques on \ac{XCT} images. 
A 2D segmentation method was employed in \cite{M3} on lab-based \ac{MCT} images of carbon fiber reinforced polymers to tackle the challenges caused by noise, low contrast between fiber and polymer, and unclear fiber gradients. A 3D instance segmentation method was developed in \cite{M4} for \ac{XCT} scans of short glass fiber reinforced polymers. The model has an additional output for embedding learning, which allows a clustering algorithm to distinguish among various fiber instances.  
Different 2D and 3D semantic segmentation techniques were applied on \ac{XCT} image data to study microstructures of materials in \cite{M5}. The ground truth information of 3D X-ray diffraction measurements was used to develop a grain-wise segmentation model for Al-Cu specimens with additional post-processing to enhance visible grain boundaries and reduce over-segmentation. A \ac{MCT} image processing method to build digital material twins was presented in \cite{M2}, where a deep learning model was applied on 2D glass and 3D carbon reinforcements' images to efficiently segment them based on extracted multi-scale features using data-driven convolutional filters. The scanned \ac{MCT} images and images produced by computer-generated virtual reinforcements models were used to train the model.

\subsection{Electronic Industry}
The electronic industry and more specifically semiconductor manufacturing has been under rapid development in the last few decades \cite{E2}, which increases the necessity of developing fast and accurate methods for defect detection \cite{E2}, unwanted particles deposition \cite{E3}, volumetric inspection \cite{E4}, etc. Therefore, the X-ray imaging technologies along with the CV techniques have drawn researchers' attention also in this field. 

In order to characterize, measure, and optimize the design and production of buried interconnects in advanced integrated circuit packages, \ac{XCT} imaging was used in \cite{E1} to avoid cross-section of the chips, and then several \ac{DL}-based 3D object detection and segmentation methods were used to identify the components and perform 3D metrology. In \cite{E3}, a \ac{CNN} model was trained on a set of energy-dispersive X-ray and scanning electron microscopy images to classify the chemical composition of particle defects on semiconductor wafers to decrease analysis time and error caused by human unpredictability.
The internal wire bonding of chips is a process that can easily face interference and produce defects in the semiconductor enterprise capsulation step. Therefore, two algorithms were used in \cite{E2} to distinguish defective chips based on the standard template and similarity calculations among the neighbor chips. Also, as ML techniques need a lot of labeled data, a data synthesis procedure was employed in \cite{E4}, where synthetic \ac{XCT} images were produced during the miniature fabrication of thin silicon wafer layers with known orientation, position, and geometry features. These known data characteristics were used as annotations and used to train an automatic ML-based feature extraction model.

\subsection{Others}
Besides the fields mentioned above, there are other industrial production fields that employ different \ac{CV} techniques on X-ray images. A brief overview is provided here. 

In \cite{O6}, a two-stage method was employed for X-ray cargo image inspection to solve the empty container verification problem. First, a rule-based algorithm was adopted to discover the location of containers' positions in the images and, afterward, a \ac{DL} method was used to identify the empty containers.
Solder balls' head-in-pillow defects were inspected using an \ac{ML}-based methodology in \cite{O10}. These defects affect the solder balls' conductivity and consequently lead to intermittent failures. In another study \cite{O02} focusing on solder joints, solder voids and head-in-pillow defects were recognized. 

In order to increase the safety in aircraft flying, a \ac{DL} method was proposed in \cite{O1} for X-ray image-based non-destructive examination of aeronautics engines with multiple defect inspection paradigms. 
Another application of using X-ray images and CV techniques is assembly inspection of internal components \cite{O14}. In this study, to ensure that all components of a complex product are assembled accurately, a multi-view X-ray imaging technique was used to obtain projection information on each internal component. Then, a deep \ac{CNN} model was used to classify the internal components and provide their coordinates to compare and match the locations and consequently recognize transposition or dislocation faults. To detect and reject defective products in a mineral wool production line, a binary classifier was developed in \cite{wool}. The goal of this work was to achieve fast classification for a real-time application that can outperform a thresholding-based method on the production line. To this end, the authors performed structured parameter pruning on the adopted deep learning model. 

X-ray-based adaptive defect detection in milled aluminum ingots surfaces was used in \cite{O2}. Automatic segmentation of multi-class progressive matrix damage of aerospace-grade advanced composite laminate images obtained by non-destructive on-site mechanical tests coupled with synchrotron radiation computed tomography was considered in \cite{O4}. In order to assess and classify tablets' internal defects, an X-ray-based method was used in \cite{O12} to explore the impacts of a filler composition, roller compaction force, and magnesium stearate on tablets quality. Also, it was shown that the use of X-ray images with quantitative \ac{CV} analysis can generate deeper mechanical knowledge of the compaction phenomenon in tableting.

A two-step CV methodology was proposed in \cite{O5} to detect voids and segment concrete samples. Another study on concrete fractures \cite{O8} noted that due to the low number of pixels for each fracture in X-ray images, high-frequency noise, and weak contrast over fractures, the performance of conventional segmentation methods is limited in extracting the continuous fractures, which leads to an overestimation of fractures aperture and thickness values. Therefore, an encoder-decoder network was adopted with a \ac{CNN} to achieve rapid and precise detection of barely seen micro-fractures. The organic microcapsules in cement were classified in \cite{O03} into five categories, namely microcapsules, ruptured microcapsules, pores, adhesive objects, and others, using a \ac{CV}-based classifier. 

\section{Computer Vision Techniques}
\label{sec:cvtechniques}

In this section, the computer vision and machine learning methods used in the previous X-ray-related studies are summarized. While advances in deep learning during the last decade have made deep learning techniques the default solutions for many machine learning tasks, these techniques require large amounts of training data, which is not always available, and therefore, traditional techniques are still commonly used in many tasks. A major difference between traditional and deep learning methods is that deep learning methods can typically operate directly on high-dimensional raw data, such as X-ray images, while traditional techniques generally use as their inputs lower-dimensional features extracted from the raw data. Therefore, feature extraction techniques are important for traditional techniques, but not for deep learning-based methods. On the other hand, as deep learning models require large training datasets, different approaches that allow training models with less data have become important. We divide our description into two main categories, traditional and deep learning methods, and further into relevant subcategories.   

\subsection{Traditional Methods}
Numerous \ac{CV} techniques have been implemented on a variety of image analysis tasks and applications. In this section, we provide a comprehensive overview of traditional (non-deep learning) \ac{CV} methods that have been applied on different X-ray image analysis tasks.

\subsubsection{Feature Extraction Techniques}
\label{sssec:features}

Traditional \ac{ML} methods typically cannot directly use the original high-dimensional raw data as their inputs, thus feature extraction methods are needed to transform the data to lower-dimensional features conserving relevant information for the analysis task at hand.

\emph{\ac{SIFT}} \cite{lowe2004sift} and \emph{\ac{SURF}} \cite{bay2006surf} are used to extract features of local image patches and have been extensively used especially in object detection applications. In \cite{bastan2011words} and \cite{S26}, \ac{SIFT} and/or \ac{SURF} were used to create \ac{BOW} \cite{csurka2004visual} representations of bag inspection images to classify them with \acp{SVM}. Well-known feature extraction methods \emph{Gabor features} \cite{lee1996gabor}, \emph{\ac{HOG}} \cite{dalal2005hog}, and \emph{\ac{LBP}} \cite{ojala2002lbp} were used in \cite{C7} to extract features for casting defect detection with several traditional classifiers. \ac{LBP} features were observed to give better results than Gabor or \ac{HOG} features. Similarly, Gabor, \ac{LBP}, \ac{HOG}, \ac{SIFT}, \ac{SURF}, and other features including features extracted from pretrained deep \acp{CNN} were used in \cite{O3} for classifying small image patches (defect vs. no defect) of casting images and the best performance was obtained with \ac{LBP} features. 

Many works applied also \emph{subspace learning} methods, most commonly \emph{\ac{PCA}}, for feature extraction. In \cite{S16}, a supervised multi-label dimensionality reduction method, Multi-Output Proximity Embedding (MOPE) \cite{tingting2012mope}, was used in feature extraction for threat classification from security images, where \ac{MOPE} is the embedding engine.

\emph{\ac{XASR}} was introduced in \cite{S24} for object recognition in security screening. It is a learning-based representation, where several patches of each object in the training set are used to learn a representative sparse dictionary for the class. In the test phase, the unseen samples are classified using these dictionaries and \ac{SRC}, which is introduced in Section \ref{sssec:classifiers}. The approach proposed in \cite{S24} led to promising results compared to other traditional feature extraction techniques.

It is also possible to use feature extraction as a preprocessing step before deep learning methods. In \cite{O2}, \ac{DOG} and \ac{MGRTS} were used for \ac{ROI} extraction in aluminum ingot images for surface defect detection. The found \acp{ROI} were subsequently classified using a \ac{CNN}.

\subsubsection{Traditional Classifiers}
\label{sssec:classifiers}

\emph{\Acp{FNN}}, \emph{\acp{ANN}}, \emph{\acp{FCNN}} and \emph{\acp{MLP}} typically refer to the same approach: a set of neurons arranged in layers and having a connection (weight) between every pair of neurons in subsequent layers. However, it should be noted that also deep learning methods are \acp{ANN} and many of them are also \acp{FNN}. The layers of traditional \acp{MLP} are commonly called fully-connected layers and used also in deep learning. The traditional \acp{MLP} are multi-input and -output functions, but due to computational limitations, the input dimension cannot be very high if the network has only fully-connected layers. During training, the network learns the connection weights so that the error between the predicted and ground-truth outputs is minimized. The error is quantified using a loss function and training happens via back-propagation of the loss so that the loss reduces.

\Acp{MLP} have been used in several X-ray analysis works including \cite{E4, W6, W13, W10, S2}.
In \cite{W6, W13, W10}, they were used for welding defect analysis, in \cite{E4} to evaluate synthesized data, and in \cite{S2} they were compared against other traditional classifier types in baggage image segmentation. 

\emph{\acf{SVM}} \cite{cortes1995svm} is a classifier that has been originally developed for binary classification and aims at finding a hyperplane that separates the features of samples in different classes by maximizing the margin between the classes. Non-linear decision boundaries can be obtained using \ac{SVM} together with the kernel trick and, e.g., by using \ac{RBF} or polynomial kernel. \Acp{SVM} can be also used for multiclass classification by formulating the problem as multiple one vs. one or one vs. all tasks. 

In X-ray image processing, \acp{SVM} have been used in \cite{O3, O10, W11, W12, W15, S2, bastan2011words, S26, S16} as the classifier. In \cite{O3}, both linear and \ac{RBF} \acp{SVM} were used for classifying patches of cast images (defect vs. no defect). In \cite{O10}, they were used to complement a \ac{CNN} in solder ball defect inspection (defect vs. no defect). Multiple works for welding detect segmentation \cite{W11, W12, W15} used first some simple methods (e.g., thresholding) for segmenting the defect candidates and then a separate \ac{SVM} to classify the candidates. Binary classification with different kernels was applied in \cite{W12, W15}, whereas \cite{W11} applied multiclass \acp{SVM} using both one vs. one or one vs. all approaches. In \cite{S2}, three \ac{SVM} models with different kernels (linear, \ac{RBF}, polynomial) were used to segment X-ray  baggage security images into organic and inorganic material. 
In \cite{bastan2011words, S26}, \acp{SVM} were used for binary classification of security images (gun or no). \ac{MOPE}-\ac{SVM} was used as the classifier in the threat classification systems in \cite{S16}. 

\emph{\Ac{RF}} applies bootstrap aggregation to build multiple classification trees and then classifies the objects based on the majority vote of the trees. \ac{RF} was compared against other traditional classifiers in \cite{S2} for an X-ray baggage security segmentation, but it showed inferior performance. In \cite{O7}, \ac{RF} was compared against k-nearest neighbors classifier in an imbalanced mineral phase segmentation task and the algorithms achieved similar performance.

\emph{\Ac{KNN}} is a simple classification technique that assigns labels for test samples by calculating the distances of the samples with all the training samples, finding k nearest samples and then selecting the most frequent label or the average of the labels in case of classification or regression, respectively. \Ac{KNN} can become noticeably slow when the data size grows. It was applied in the same studies as \ac{RF} \cite{S2, O7} with similar results.

\emph{\Ac{NB}} is a simple Bayes theorem-based probabilistic classifier with (naive) independence assumption among the features. It was the winner among the traditional classifiers compared for X-ray baggage security segmentation in \cite{S2}.

\emph{Logistic regression} is used to evaluate class probabilities in binary classification tasks. It aims at finding optimal parameters values to fit a logistic function to model a binary target variable. The parameters are usually estimated using maximum likelihood estimation over cross-entropy loss. Logistic regression was applied in \cite{O5} to differentiate the features of aggregate and mortar pixels in concrete phase segmentation. 

\emph{\Acf{SRC}} \cite{wright2009src} computes a sparse representation for all training samples and decides the class of an unseen test sample by evaluating how well the sample can be constructed from the sparse representations of different classes. The assumption is that the sparse representations capture the central features of the images belonging to a certain class, and a better reconstruction means that the central features of a test sample follow the class characteristics. In \cite{S24}, \ac{SRC} was used together with the \ac{XASR} representations introduced in Section \ref{sssec:features} for security screening. 

\subsubsection{Clustering Techniques}

Clustering refers to a process of grouping the input samples so that similar items are assigned to the same cluster and dissimilar ones into different clusters. Segmentation can be seen as a clustering task, where regions corresponding to different objects should be assigned into different clusters and evaluated via clustering metrics as described in Section \ref{ssec:clusteringindices}. Clustering techniques also find use in X-ray image segmentation. 

\emph{K-means clustering algorithm} initially assigns all items randomly in one of K clusters. Then the algorithm proceeds iteratively by computing the centroid of each cluster, reassigning the items to the cluster of the closest centroid, and repeating these steps until the algorithm converges. K-means was assessed in \cite{O7} for 3D mineral phase segmentation based on voxels' gray-scale values. 
\emph{\Ac{FCM}} is a fuzzy version on K-means. While K-means assigns each item into one cluster, in \ac{FCM} the items can belong to multiple clusters in a fuzzy manner. \Ac{FCM} was used for 3D mineral phase segmentation based in \cite{O7}.

\emph{\Ac{SLIC}} \cite{achanta2012slic} is another variant of K-means, where the distance measure combines feature similarity and spatial distance of the pixels. Also, the number of distance evaluations is limited to an area proportional to the superpixel size. \Ac{SLIC} was applied for sub-component level segmentation in anomaly detection within X-ray security imagery in \cite{S11}.

\subsection{Deep Learning Methods}

Deep learning has dominated many \ac{CV} tasks by adopting deeper and more complicated neural architectures that make the networks capable of modeling more complex patterns and relations. X-ray image processing is not an exception and researchers have exploited the benefits of \ac{DL} models, too. In this section, we present a comprehensive overview of \ac{DL} methods applied on X-ray image analysis. We review the adopted deep architectures categorized according to the considered CV tasks, i.e., classification, detection, and segmentation, and we also cover different loss functions, data augmentation strategies, and other approaches for improving the performance of deep learning models.

A type of deep learning model that has been commonly used for image analysis is \acfp{CNN} \cite{raitoharju2022convolutional} leading to high performances \cite{O10, W4}. \ac{CNN} architectures consist of several layers with different properties. The most common types of layers are: convolutional layer, pooling layer, and fully-connected layer. Convolutional layers learn to extract useful features from the input images and each layer transforms the input data into a more abstract representation. Pooling layers are used to compress the feature maps and fully-connected layers make the final prediction based on the extracted features. The last layer provides the network output and the output format depends on the \ac{CV} task at hand.

\subsubsection{Deep Classification and Backbone Architectures}

In \ac{CNN} architectures for classification, the output format usually is a one-hot encoded vector \cite{raitoharju2022convolutional}, which has one element for each class and the value of the elements is a form of predicted probability of the input image to belong to the corresponding class. The image is assigned to the class with the highest probability. This type of layer is typically used together with categorical cross entropy loss function (see Section~\ref{sssec:lossfunctions}) and softmax activation function defined as 
\begin{equation}
    \hat{\mathbf{y}}[i] = \frac{e^{\mathbf{y}[i]}}{\sum_{c=1}^C e^{\mathbf{y}[c]}},
    \label{eq:softmax}
\end{equation}
where $\mathbf{y}[i]$ and $\hat{\mathbf{y}}[i]$ denote the $i^{th}$ element in the output vector before and after applying the softmax function, respectively, and $C$ is the number of output classes. Softmax activation confines the output element values between zero and one and makes the sum of the output elements equal to one. The exponential function highlights the probability of the most probable class making the predictions clearer.

While the \ac{CNN} architectures presented below were originally proposed for classification tasks, they can be used as backbone networks in other problems, such as object detection \cite{C3}, or within \acp{GAN} used for data augmentation \cite{C9} simply by removing the output layer designed for classification and adding other types of layers.

\begin{figure*}[!t]
\centering
\includegraphics[width=\textwidth]{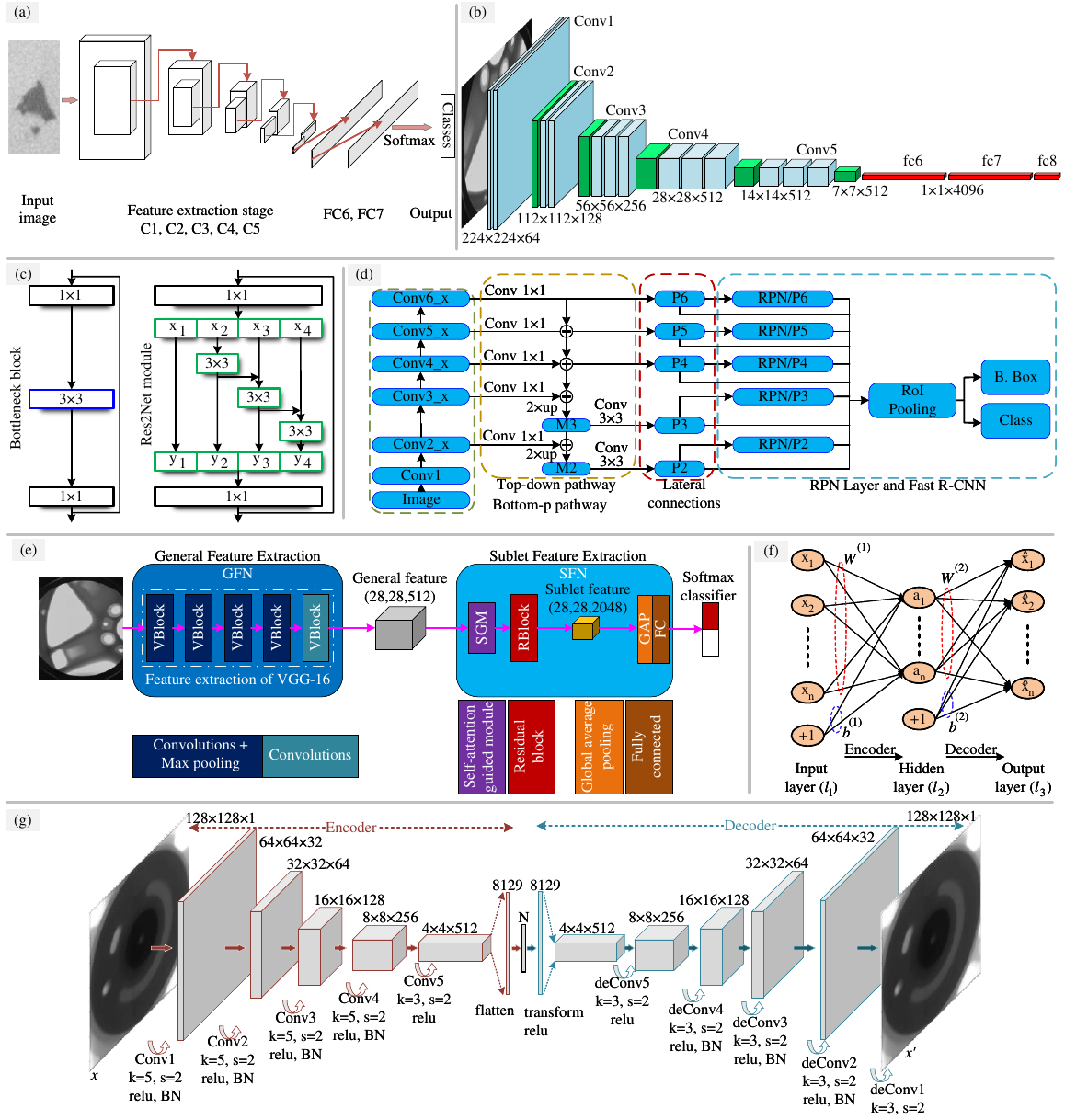}
\caption{Deep classification architectures: a) ED-CNN, b) VGG-16, c) ResNet vs. Res2Net, d) DetNet-59, e) Self-attention guided CNN, f) \Acl  {CAE}, and g) \Acl{SAE}.}
\label{DeepClassificationArchitecturs}
\end{figure*}

Studies applying simple \ac{CNN} architectures for X-ray image classification include \cite{W03} using a simple \ac{CNN} model for weld defect classification, \cite{O03} using another simple \ac{CNN} model for binary classification of automotive components into defective and non-defective class, and \cite{C2} using \emph{Evenly Distributed CNN (ED-CNN)} illustrated structure in Fig. \ref{DeepClassificationArchitecturs}-a for casting defects classification.

A simple and well-known classification backbone architecture is 
\emph{VGG} \cite{VGG} which contains stacked convolutional and max pooling layers. VGG-16 architecture shown in Fig. \ref{DeepClassificationArchitecturs}-b was used in \cite{O02} for 
solder joint classification
and in \cite{C06} along with spatial attention and bilinear pooling for casting defects classification. VGG-16 was used as a feature map extractor (backbone network) in an anomaly detector network for casting defect localization in \cite{C9} and in an object detection network for X-ray baggage security assessment in \cite{S9}. In \cite{W01}, a different VGG variant was used for weld defect classification. Different simple \ac{CNN} models including VGG-19, VGG-F, and VGG-2048 were compared in \cite{O3} for casting defects classification from X-ray data.

One of the most common and well-known classification architectures is \emph{Residual neural network (ResNet)}. ResNet is an extension on \ac{CNN} models that was proposed to prevent the problems caused by very deep networks, in particular the vanishing gradient problem. To optimize and overcome the network degradation problem, ResNets include residual blocks that have skip connections over some layers (Fig. \ref{DeepClassificationArchitecturs}-c, left). They also apply batch normalization after each convolution layer. These factors make it easier to pass information through the networks, which allowed to have a larger number of layers and a smaller error rate on both train and test sets than the earlier \ac{CNN} models. Due to its benefits, different variants of ResNet with various depths, such as ResNet-18 \cite{M2}, ResNet-34 \cite{O6}, ResNet-50 \cite{S7}, and ResNet-101 \cite{C9}, have been used in X-ray data assessment studies, e.g., in casting defect recognition \cite{C9, C04} and detection of internal defects in the Aluminum Conductor Composite Core (ACCC) \cite{M04}. 

In order to improve the multi-scale performance of ResNet, a newer model known as \emph{Res2Net} was introduced in \cite{Res2}. In Res2Net, the residual blocks are replaced by hierarchical residual-like connections within one single residual block (Fig. \ref{DeepClassificationArchitecturs}-c, right). Res2Net was used as the generator in a \ac{GAN} in \cite{S1} for data augmentation (See  \ref{sssec:augmentation}). A modification of ResNet-50, \emph{DetNet}, was proposed in \cite{detnet} as a backbone network optimized for object detection alleviating the loss of location information in feature maps caused by down-sampling operations. DetNet-59 architecture shown in Fig. \ref{DeepClassificationArchitecturs}-d was adopted in \cite{C3} as a backbone for casting defect detection.

While convolutions focus on local relations in the data, augmenting convolutions with different approaches to capture
long-range dependencies have been proposed. Self-attention \cite{zhao2020exploring} is an attention mechanism that can relate different positions
of the data in order to compute a feature representation. A \emph{self-attention guided CNN} was used in \cite{C5} to detect small casting defects. Its overall structure of the employed model is shown in Fig. \ref{DeepClassificationArchitecturs}-e.

Considering the special characteristics of the weld defects, the usual pooling strategies have poor dynamic adaptability. Therefore, an improved pooling strategy was proposed in \cite{W04}. In the proposed approach, different pooling method were used depending on whether the pooling domain is outside the defected area or on the defect's edge.

Autoencoders are a special type of neural networks that can be used for unsupervised feature extraction. They are composed of an encoder that turns input images into feature representations and a decoder that tries to reconstruct the input from the feature representation as the network's output. While the task would be trivial if the feature representation had the same dimensionality as the input (and output), the feature representation in autoencoders usually has a much lower dimensionality, which forces the network to learn representations that contain the most useful information for the reconstruction process. As the training requires only the images with no need for class labels, training can be performed in a fully unsupervised manner. The trained encoder can be then used as a feature extractor for other tasks, such as classification. This approach can be useful when there is a large unlabeled data set available, but only a limited number of labeled training samples. For instance, a non-convolutional \emph{\ac{SAE}} shown in Fig. \ref{DeepClassificationArchitecturs}-f was utilized in \cite{W7} as an intrinsic feature extractor for welding defect detection. In \cite{C03}, an unsupervised inspection system was built on top of a \emph{\ac{CAE}} (Fig. \ref{DeepClassificationArchitecturs}-g) to inspect casting X-ray images with no labeling. In \cite{A5}, a \ac{CAE} was trained using abundant normal images of manufacturing production lines. The encoder was then combined with fully-connected layers for classification that were trained using a lower number of labeled samples of both normal and defective engines.

\subsubsection{Deep Object Detection Architectures}
\label{sssec:detectionarchitectures}

Object detection networks aim at finding the locations of objects in addition to recognizing them. The number of objects can significantly vary in different images, which means that the output layers for object detection architectures cannot use a fixed fully-connected structure as is commonly done in classification. The key design question in object detection architectures is how to locate the possible objects for deeper analysis. One approach would be to predefine all possible bounding box locations and sizes and exhaustively analyze whether they contain objects of interest. However, this approach would have an enormous number of bounding boxes to analyze and would be computationally too expensive. Therefore, most object detection architectures propose approaches for finding only the most promising subset of all the possible bounding boxes for further analysis.

\begin{figure*}[!t]
\centering
\includegraphics[width=\textwidth]{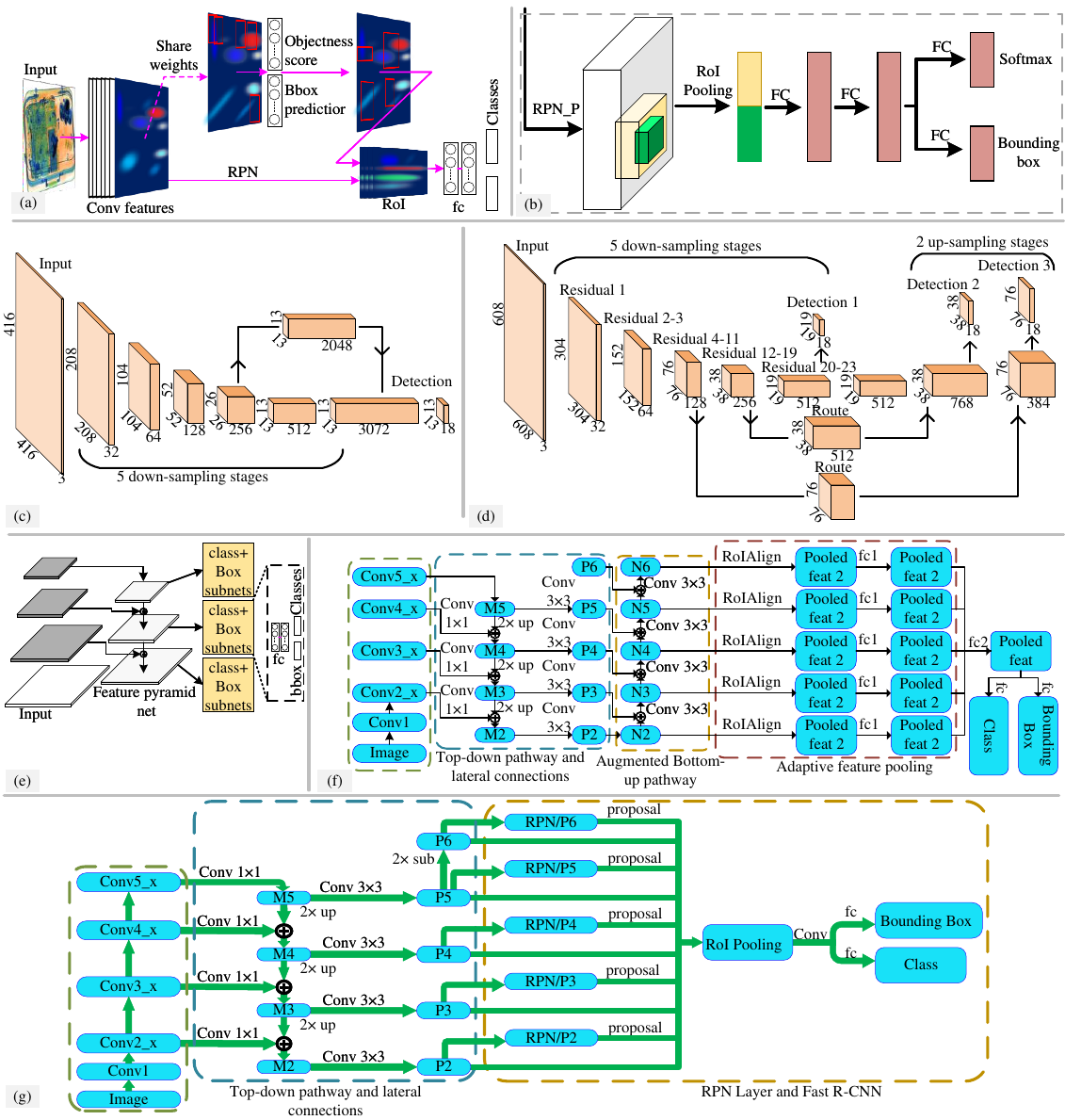}
\caption{Deep detection architectures: a) Faster R-CNN, b) PDN branch, c) YOLOv2, d) YOLOv3, e) RetinaNet, f) PANet with ResNet-50, and g) FPN.}
\label{DeepDetectionArchitecturs}
\end{figure*}

\emph{Region-based CNN (R-CNN)} \cite{girshick2014rich} is one of the architectures commonly used for object detection. For a given image, R-CNN applies a selective search mechanism to extract approximately 2,000 \acp{ROI}. Afterward, each \ac{ROI} is introduced to a \ac{CNN} to obtain the output features, and then a collection of SVM classifiers is used to recognize the type of object in the \ac{ROI} (if there is any).
\emph{Fast R-CNN} \cite{girshick2015fast} improves the efficiency of R-CNN by not introducing all \acp{ROI} to the \ac{CNN}, but introduces the input image once, and the features for the \acp{ROI} are then extracted from the overall feature map. An upgraded version of Fast R-CNN, called \emph{Faster R-CNN} (Fig. \ref{DeepDetectionArchitecturs}-a) uses a separate network to predict \acp{ROI} instead of using the slow selective search. Faster R-CNN was applied in \cite{O1} and \cite{O13, S9, S10, S13} to detect defects in tires and prohibited items in baggage, respectively. In \cite{S14}, an additional branch called \emph{Part-based Detection Network (PDN)} was added to Faster R-CNN (Fig. \ref{DeepDetectionArchitecturs}-b) to improve detection of occluded items in threat detection on X-ray security images.

\emph{You Only Look Once (YOLO)} \cite{yolo} is another commonly-used object detection architecture. The main difference between YOLO and the region-based approaches is that YOLO uses a single \ac{CNN} to predict both bounding boxes and class probabilities. Therefore, it can be trained in an end-to-end manner and it is much faster than the region-based approaches. There are multiple versions of YOLO architectures and many of them have been employed also in X-ray image processing. \emph{YOLOv2} \cite{yolov2} improves the original YOLO in multiple ways, such as adding batch normalization, removing fully-connected layers, and using anchor boxes. As a result, YOLOv2 manages to improve YOLO's recall and localization, while maintaining its classification accuracy. YOLOv2 was used in \cite{M1} with the architecture shown in Fig. \ref{DeepDetectionArchitecturs}-c for detection and classification of diffraction patterns in single-particle imaging. \emph{YOLOv3} \cite{yolov3} improves the accuracy of earlier versions by adding objectness scores to bounding box prediction, adding connections to the backbone network layers, and making predictions at three separate levels of granularity to improve performance on smaller objects. YOLOv3 was employed in \cite{M1} as a diffraction pattern detector on X-ray images by the illustrated structure in Fig. \ref{DeepDetectionArchitecturs}-d and in \cite{S18, isaac2021multi} to detect dangerous objects in baggage security application. It also was used in \cite{xue2022high} to detect defects in casting products. To achieve better detection speed and accuracy, \emph{YOLOv4} is introduced in \cite{bochkovskiy2020yolov4} with improvements in network structure, training method, loss function, and data enhancement in comparison to YOLOv3. YOLOv4 was used in \cite{zhou2021x} as a detector in an X-ray security inspection task. The fifth version of YOLO, known as \emph{YOLOv5} was used in \cite{C05} to detect casting defects.

\emph{Single Shot Multibox Detector (SSD)} \cite{ssd} is a one-stage object detection network that eliminates the proposal generation phase by discretizing the bounding box prediction space into a set of default boxes and then calculating scores presenting the existence of each object class in each box and finally makes adjustments on the boxes to improve the scores. During inference, the predictions obtained from multiple feature maps with different resolutions are combined together to capture various object sizes. This method is combined with VGG-16 and ResNet-101 in \cite{C8} as an object detector for casting assessment.

Another common one-stage object detection architecture, \emph{RetinaNet} \cite{Retina}, uses focal loss and feature pyramid network (Fig. \ref{DeepDetectionArchitecturs}-e). It achieves good performance with dense and small-scale objects, as the focal loss better addresses the problems caused by a major class imbalance between background and foreground classes. In the topic of X-ray image analysis, RetinaNet is used in \cite{W2, S10, S12} to detect defects in welding, anomalies in cluttered security imagery, and firearms in baggage security imagery, respectively.

In \cite{S3}, a \emph{\ac{CST}} framework for detection and classification of heavily occluded baggage items from X-ray scans was proposed. The framework uses non-convolutional \ac{CST} approach for object proposal extraction and a \ac{CNN} only for subsequent object recognition.

\emph{\ac{PANET}} \cite{panet} can be used for both detection and segmentation tasks. It improves region-based networks by including bottom-up path augmentation to cut down the information path among lower layers and topmost features, adaptive feature pooling to connect feature grids at all levels of features, and fully-connected fusion to enhance  mask prediction. It was combined with Resnet-50 in \cite{C3} (shown in Fig. \ref{DeepDetectionArchitecturs}-f) to detect defects in casting products.

\emph{\acp{FPN}} \cite{fpn} were introduced for detecting objects at different scales. They use the inherent pyramidal and multi-scale hierarchy of deep \acp{CNN} to build feature pyramids with marginal extra cost. A \ac{FPN} was used for detection of automobile casting aluminum parts in \cite{C6} with the shown structure in Fig. \ref{DeepDetectionArchitecturs}-g. In \cite{S7}, it was combined with a ResNet-50 and used for prohibited item detection in X-ray scanning images.

\subsubsection{Deep Image Segmentation Architectures}
\label{sssec:segmentationarc}

\begin{figure*}[htbp]
\centering
\includegraphics[width=0.9\textwidth]{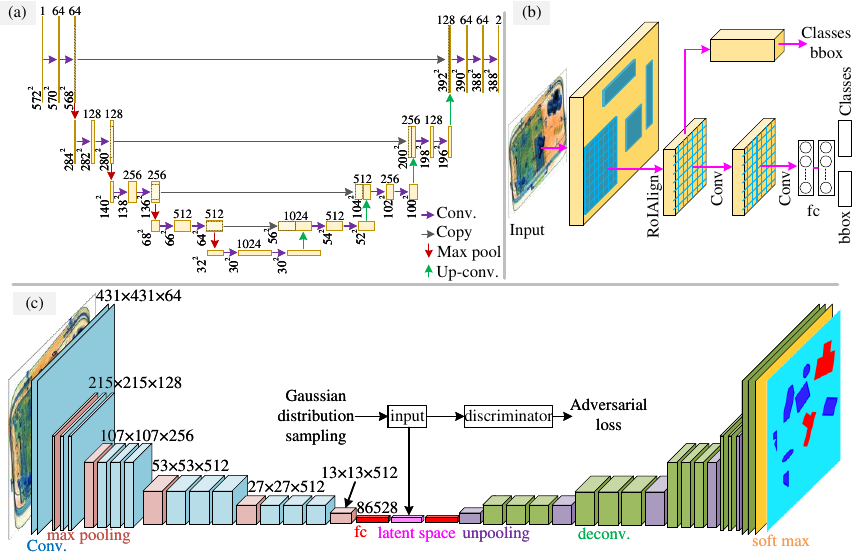}
\caption{Deep segmentation architectures: a) U-Net, b) Mask R-CNN, and c) CHNet.}
\label{DeepSegmentationArchitecturs}
\end{figure*}

In image segmentation, the network output is an image with the same dimensions as the input image, and the task is to predict the class of each pixel. Segmentation architectures typically have a structure similar to autoencoders, where an encoder learns to extract a descriptive lower-dimensional representation of the input and a decoder up-samples the feature map size back to the input size and produces the class predictions. The networks are often fully-convolutional networks, i.e., they do not contain any fully-connected layers.

\emph{U-Net} \cite{ronneberger2015unet} (Fig. \ref{DeepSegmentationArchitecturs}-a) is one of the widely used models for image segmentation. It has an encoder-decoder structure with additional connections between them. The method has been used on both 2D and 3D (an extension upon the standard U-Net) X-ray data in \cite{A1, M3, M5}. In \cite{A1}, defects in 3D AM X-ray images were segmented using a modified U-Net. U-Net was used to segment continuous carbon fiber reinforcements composites in \cite{M3}. 2D and 3D U-Net were applied in \cite{M5} to predict grain boundaries in Al-Cu alloy materials. U-net structure with additional skip connections was used in \cite{W1} for segmenting the locations of welding defects. A U-Net-like structure with residual connections was used in \cite{A3} for porosity segmentation in \ac{XCT} scans of additively manufactured metal specimens.

\emph{Mask R-CNN} \cite{he2017mask} is an extension of Faster R-CNN (see Section~\ref{sssec:detectionarchitectures}) with an additional output for predicting segmentation masks for each \ac{ROI}. As this approach segments each instance of an object class independently, Mask R-CNN is an architecture for \textbf{instance segmentation}. In \cite{S10}, Mask R-CNN (Fig. \ref{DeepSegmentationArchitecturs}-b) was used to segment anomalies in cluttered security imagery. Casting defects and firearms in baggage were segmented using Mask R-CNN in \cite{C8, S12}.

\emph{CH-Net} \cite{S4} is a semantic segmentation model based on adversarial autoencoders (AAEs) \cite{aae} (see Fig. \ref{DeepSegmentationArchitecturs}-c). It was proposed as a fast and memory-efficient method for baggage security image segmentation in \cite{S4}. \emph{DeepLabv3+}, an encoder-decoder network with atrous separable convolutions in the encoder was proposed in \cite{Chen_2018_ECCV}. DeepLabv3+ was applied with ResNet18 as a backbone for semantic segmentation of \ac{MCT} images for creating digital material twins of fibrous reinforcements in \cite{M2}.

\subsubsection{Loss Functions}
\label{sssec:lossfunctions}

As quantifiers of the difference between a model's predicted and expected outcomes, loss functions are essential in training deep neural networks. During training, the models are guided towards minimizing this difference and, if the loss function does not represent well the problem at hand, the results will be suboptimal.

In regression tasks, where the goal is to learn to predict specific values as the network's output, \emph{\ac{MSE}} is a common loss function. \ac{MSE} loss is defined as 
\begin{equation}
    L(\hat{\mathbf{y}}_s, \mathbf{y}_s)  = \frac{\sum_{i=1}^{N} \left(\hat{\mathbf{y}}_s[i]-\mathbf{y_s[i]}\right)^2}{N},
    \label{eq:mse}
\end{equation}
where $\mathbf{y}_s$ is a vector containing the target values for all the network's $N$ outputs for a specific sample $s$, $\hat{\mathbf{y}}_s$ is the predicted output vector, and $\mathbf{y_s[i]}$ is the $i^{th}$ element of the target output. The final loss is the average loss over the training samples.

While \ac{MSE} can be also used in classification tasks, \emph{cross entropy} loss, also called \emph{log loss}, is preferred in classification tasks in general, and also most of the works on X-ray image classification use cross entropy loss. There are two commonly used versions of the cross entropy loss. \emph{Binary cross entropy} loss is used in binary classification tasks as well as in multi-class multi-label classification tasks, where each sample may belong to multiple classes:
\begin{equation}
    L(\hat{\mathbf{y}}_s, \mathbf{y}_s)  =  -\sum_{i=1}^{N} \left(\mathbf{y}_s[i] \log( \hat{\mathbf{y}}_s[i]) +(\mathbf{y}_s[i]) \log(1- \hat{\mathbf{y}}_s[i])\right),
\end{equation}
\emph{Categorical cross entropy} loss is used in multi-class single-label classification tasks:
\begin{equation}
    L(\hat{\mathbf{y}}_s, \mathbf{y}_s)  = -\sum_{i=1}^{N}\mathbf{y}_s[i] \log(\hat{\mathbf{y}}_s[i]).
\end{equation}
Compared to \ac{MSE}, cross entropy losses penalize output values that lead to wrong classification more. Similar to \ac{MSE} loss, the binary cross entropy loss gives an equal weight for all the output elements, whereas the categorical cross entropy loss focuses on positive samples, i.e., learning which samples should be classified to a specific class instead of trying to learn which samples should not be classified to the class \cite{raitoharju2022convolutional}.

In object detection tasks, the models need to predict the bounding box locations along with the corresponding class. Thus, they usually minimize both a regression loss (\ac{MSE}) to learn the bounding box locations and a classification loss (cross entropy) to learn the classes. As semantic segmentation can be seen as pixel-wise classification, pixel-wise cross entropy loss is commonly used also in segmentation tasks. Below we briefly introduce some less common loss function choices that have been considered in X-ray image analysis tasks.

\emph{Triplet loss} \cite{triplet_loss} is a loss function that tries to reduce the distance between data in the same class and increase the distance between data belonging to different classes. To compare performance, the triplet loss with cosine similarity is used as the loss function in \cite{O10} for soldering defect inspection, leading to a higher accuracy compared to using the cross-entropy loss function.

\emph{Mutual-channel loss} function was introduced in \cite{MC-loss} and it consists of a discriminant component and a diversity component. This results in a set of feature channels each of which reflects different locally discriminative regions for a particular class. This loss function was used for casting defect detection in \cite{C1} to focus on different discriminative regions without part annotations or bounding boxes of the defects.

\emph{Focal loss} was designed to tackle multi-class object detection scenarios with a high imbalance between foreground and background classes \cite{focal_loss}. It gives a higher weight to hard misclassified examples. It was used in \cite{O2} for an imbalanced dataset of milled aluminum ingot defects.

Although focal loss makes one-stage detectors focus more on hard samples for improving performance, the availability of a fair amount of hard outlier samples can cause a reduction in accuracy \cite{zhou2021x}. \emph{Gradient Harmonization Mechanism (GHM) loss} was introduced in \cite{li2019gradient} to tackle this problem and it was used in \cite{zhou2021x} on a YOLOv4 model in security threat detection application.

\subsubsection{Data Augmentation}
\label{sssec:augmentation}

Data augmentation refers to the process used to increase the number of data by creating slightly modified versions of the available real data or creating synthetic data. It can help to reduce the overfitting problem. Common augmentation tricks, such as random rotation, cropping, or flipping, are often used for training deep learning models. Some more specific augmentation techniques that were used in X-ray related studies are presented here.

A common way to tackle the problem of lacking a high number of images from real-world environments is \emph{learning based image synthesis}. However, these methods usually combine background and foreground images randomly which limits the performance of the generated data. In X-ray security applications, a learning-based image synthesis method was proposed in \cite{W08}. In this method, a detector is first trained to estimate difficult positions for each foreground object detection. Then, a so-called difficulty map is created and the objects are synthesized at hard-to-detect locations using the difficulty map.

\emph{Attention-Guided Data Augmentation (AGDA)}, proposed in \cite{C1}, creates new training samples from existing ones by suppressing the most discriminative parts found using attention maps. The method was applied for casting defect detection in~\cite{C1}.  

\begin{figure}[!t]
\centering
\includegraphics[width=\columnwidth]{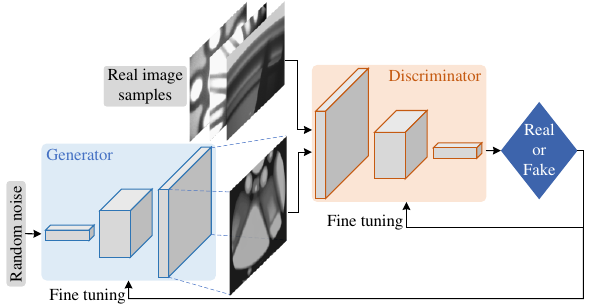}
\caption{Basic GAN structure.}
\label{gan_img}
\end{figure}

\emph{\Acfp{GAN}} \cite{GAN} are a class of \ac{DL} models that learn to generate new data samples with the same statistics as those in the training set. \Acp{GAN} consist of two competing networks: a generator that generates fake images and a discriminator that tries to distinguish between real and fake images as shown in Fig. \ref{gan_img}. To be able to fool the discriminator, the generator needs to learn to create realistic images. \Acp{GAN} were used in \cite{C01} to generate simulated data of defective aluminum casting and improve the balance of the dataset. A GAN-based method was used in \cite{S1} to synthesize X-ray security images. Several modifications of the original \acp{GAN} have been also used for data augmentation in X-ray applications. The authors of \cite{W4} noted that generating data using prior human knowledge is not applicable for some specific types of welding defects (e.g., burn through and crack of weld) due to their complexity. Therefore, they used \emph{\acp{WGAN}} \cite{WGAN} for the task. In \Acp{WGAN}, the discriminator gives a fakeness score for the generated samples instead of just classifying them into real and fake classes. \emph{Self-Attention Generative Adversarial Network (SAGAN)} was used in \cite{S5} that first generates several images of prohibited items. Subsequently, the images are transformed to X-ray format using a cycle GAN and combined to different backgrounds. Also, \emph{Deep Convolutional GAN (DCGAN)} and \emph{Spatial-and-Channel Attention Block and X-ray Wasserstein GAN Gradient Penalty (SCAB-XWGAN-GP)} were used in \cite{dumagpi2021evaluating,liu2022data} for data augmentation in X-ray security application.

\section{Evaluation Setup and Metrics}
\label{sec:evaluation}

In order to evaluate the performance of new methods and to have a valid comparison among different algorithms, it is important to know the different evaluation protocols that have been used in previous studies. Commonly, CV and ML techniques are evaluated by dividing the datasets into non-overlapping training and test sets in order to reliably estimate the performance on unseen data. If the model to be trained has some hyperparameters which need to be determined by the user, they are typically set using a third separate part of the dataset called validation set. When no predefined validation set is given by the experimental protocol defined by the dataset, two approaches can be used. In the first one, the training set is divided into two non-overlapping sets, one used for training and the second used for validation. The second approach divides the training set into $k$ non-overlapping subsets (sometimes called folds) and performs training and validation $k$ times. Each time, data in a different fold is used for validation, while the data in the remaining $k-1$ folds are used for training the models obtained by using different hyperparameter values. The average performance on all folds is calculated and the best hyperparameter values are those used in the model leading to the highest average validation performance.  Then, the final model can be trained on the full training set using the best hyperparameter values. This procedure is commonly known as k-fold cross-validation. Especially for smaller datasets, an approach similar to k-fold cross-validation approach may be used for the testing as well to get a more reliable performance estimate. In this case, the fold set aside should not be used for adjusting the model's hyperparameters, but the goal is to evaluate the method on a wider variety of test samples not seen during the training process. It can be also necessary to repeat the overall experiment multiple times and use average values, if variations in the methods' performance are expected.  

In addition to the training setup, an important aspect of the evaluation is the selection of the evaluation metrics. Accepted and standard evaluation metrics must be employed to ensure fair comparisons. For meaningful evaluations, it is also important to understand which evaluation metrics are suitable for the task at hand. Below we introduce the most commonly used evaluation metrics for classification, detection, segmentation, and speed comparisons.

\subsection{Classification Metrics}
\begin{figure}[!t]
\centering
\includegraphics[scale=.8]{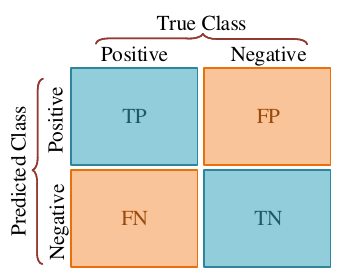}
\caption{Confusion matrix.}
\label{Confusion-matrix}
\end{figure}

Many binary classification metrics rely on the counts of True Positives (TP), True Negatives (TN), False Positives (FP), and False Negatives (FN) that are defined as shown in the confusion matrix in Fig. \ref{Confusion-matrix}. In multi-class classification, similar confusion matrices can be computed either considering all classes or independently for each class and generalized evaluation metrics can be defined using these class-specific numbers.

\subsubsection{Accuracy} 
Accuracy is maybe the most commonly used classification metric and it is defined simply as the ratio of correct classifications to the total number of samples. In binary classification, this can be defined as follows:
\begin{equation}
\eta=\frac{TP+TN}
 {TP+TN+FP+FN}.
\end{equation}
While accuracy is an intuitive measure, it is not a good evaluation metric when the class distribution is not balanced. For example, if there are dangerous items in every 1000$^{th}$ bag, a classification method can achieve a very high accuracy of 0.999 by simply classifying everything as safe. However, in such applications missing a dangerous item is obviously much more critical than incorrectly labeling a safe bag as dangerous. The latter situation might only lead to an additional manual inspection while the former may cause significant danger. Therefore, other evaluation metrics should be used in such imbalanced cases. Also, several works on X-ray image classification have complemented accuracy with other classification metrics. The combination of accuracy, \emph{recall}, \emph{precision}, and \emph{F1 score} (with possibly other metrics) has been used, e.g., in X-ray-based casting defect classification \cite{C1, C5}, welding defect recognition \cite{W7}, and concrete phase segmentation~\cite{O5}.

\subsubsection{Recall} 
Recall or \emph{sensitivity} or \emph{\ac{TPR}} is a commonly used metric for binary classification
to quantify the number of correctly identified positive samples out of all positive samples in the data. It is defined as follows:
\begin{equation}
Recall=\frac{TP}
 {TP+FN}.
\end{equation}
Recall can be a good evaluation metric, when it is important to correctly recognize samples of a particular class, while it is less critical to incorrectly label objects from the other class, as in the example of finding bags containing dangerous items. However, a perfect recall can be always obtained simply by labeling all the samples as positive. Therefore, recall alone is not a sufficient evaluation metric.

\subsubsection{Precision} 
Precision is another commonly used metric for binary classification that quantifies the ratio of correctly predicted positive predictions to the total number of positive predictions. It is defined as:
\begin{equation}
Precision=\frac{TP}{TP+FP}.
\end{equation}
A good precision can be generally achieved by labeling only very few and certain cases as positive and, therefore, it is not a good measure by itself either. However, recall and precision complement each other and other metrics combining the two have been suggested.

\subsubsection{F1 Score} 

F1 score is a widely used metric for binary classification that takes both recall and precision into account and can be seen as their harmonic average. F1 is defined as:
\begin{equation}
F1=2*\frac{precision*recall} {precision+recall} = \frac{2TP}{2TP+FP+FN}.
\end{equation}
F1 score is typically considered a good single measure for binary classification and more suitable for unbalanced data than accuracy. However, it is not as intuitive to understand what a certain score means in practice.

\subsubsection{Specificity} 
Specificity or \emph{\ac{TNR}} is another metric for binary classification that can be used to complement recall/sensitivity. While recall focuses on the positive items, specificity focuses on the negative items. If the class assignment to positive and negative classes is reversed, specificity becomes equal to recall before the reversion. Perfect specificity can be obtained by classifying all the samples as negative. It is defined as
\begin{equation}
Specificity=\frac{TN}{TN+FP}.
\end{equation}
Specificity was used to evaluate binary pixel-wise welding defect classification in \cite{W1}.

\subsubsection{Average Performance} 
Average performance is another performance score that considers both precision and recall defined in \cite{S23}. This score is averaging the performance of the method over all available classes and it is defined as:
\begin{equation}
p=\frac{1}{N_{classes}}\sum_{i}^{N_{classes}}\sqrt{precision_i*recall_i} \:.
\end{equation}
The metric was used in \cite{S23} for evaluating dangerous item classification.

\subsubsection{\ac{MCC}}
\ac{MCC} can also represent an overall classification performance as a single value and it is considered to be a reliable metric also when the class distribution is very imbalanced. This metric is defined as follows:
\begin{small}
\begin{equation}
MCC=\frac{TP*TN-FP*FN}
 {\sqrt{(TP+FP)(TP+FN)(TN+FP)(TN+FN)}} .
\end{equation}
\end{small}
\ac{MCC} was used to measure welding defect classification performance in \cite{W10}.

\subsubsection{\ac{ROC}}
\begin{figure}[!t]
\centering
\includegraphics[width=\columnwidth]{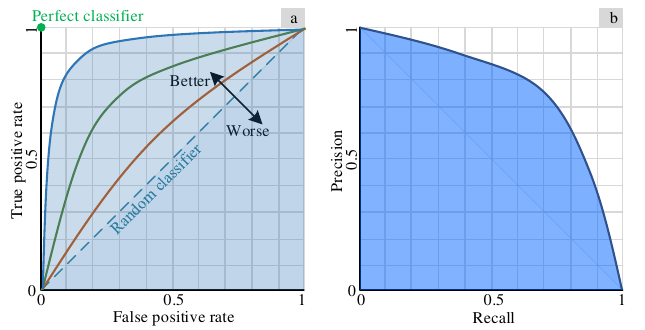}
\caption{a) Receiver Operator Characteristics (ROC) curve, b)  precision vs. recall curve.}
\label{ROC-curve}
\end{figure}

\ac{ROC} curve and \emph{ Precision-Recall curve} can be used to evaluate binary classification methods when the result depends on a threshold. As explained above, perfect recall/sensitivity can be obtained by classifying all the samples as positive, while classifying all the samples as negative leads to perfect specificity or precision. When the threshold is varied so that the number of positive assignments grows from zero, a good classifier assigns the true positive items as positive before falsely assigning any negative item as positive. A method's ability to do so can be evaluated using \ac{ROC} curves that plot true positive rates vs. false positive rates (=1-true negative rate) during the process as shown in Fig. \ref{ROC-curve}-a or precision-recall curves that plot precision vs. recall pairs (Fig. \ref{ROC-curve}-b) as the name suggests. In general, \ac{ROC} curves are more suitable for balanced class distributions, while precision-recall curves are recommended for imbalanced cases \cite{davis2006curves}. 
\ac{ROC} curves were used to evaluate firearm recognition in \cite{S26}, while
both \ac{ROC} and Precision-Recall curves were used in \cite{S3} to compare algorithms for identifying normal and suspicious items.

\subsubsection{\ac{AUC}} 
\ac{AUC} allows representing \ac{ROC} (or Precision-Recall) curves numerically by computing the relative area below the curve. A higher area corresponds to better performance. \ac{AUC} score for \ac{ROC} curves was used in \cite{S3} for the detection of different suspicious items in a security application. 


\subsubsection{\ac{MAP} - classification}
\label{ssec:map}
\ac{MAP} can be used as a classification metric for methods that rank the test items based on their estimated probability to contain certain objects. Each class-specific ranked list is used to compute precision and recall values for each rank. Here, recall is defined as the ratio of positive examples ranked above a given rank and precision is the ratio of all samples above that rank that are from the positive class \cite{everingham2010pascal}. This creates a precision-recall curve. \ac{AP} for the class is defined as the average of precision values at eleven equally spaced recall levels [0,0.1,...,1]. Finally, \ac{MAP} is the average over classes. As described in Section \ref{ssec:sixray}, \ac{MAP} was adopted with SIXray dataset \cite{sixray} for the image-level classification task. In \cite{bastan2011words}, \ac{AP} is used as an evaluation metric in a binary handgun recognition task.

\subsection{Segmentation Metrics}
Segmentation can be seen as pixel-wise classification and, therefore, classification metrics can be used also as segmentation metrics. This has been common also for X-ray segmentation tasks. For example, sensitivity, specificity, accuracy, Precision-Recall curves, and \ac{AUC} on Precision-Recall curves were used to evaluate segmentation performance in welding defect localization in \cite{W1}. In \cite{A3}, Precision, Recall, and \ac{MCC} were used to evaluate AM porosity segmentation performance. 
In \cite{O7}, \ac{TPR}, \ac{FPR}, and \ac{AUC} for \ac{ROC} were used to evaluate the performance in an imbalanced mineral phase segmentation task. Accuracy, Recall, Precision, F1 Score, \ac{ROC} curve, and \ac{AUC} on \ac{ROC} were used to evaluate concrete segmentation in \cite{O5}.

\subsubsection{Dice Coefficient} 
Dice Coefficient can be used to evaluate the ground-truth segmentation mask $A$ with the predicted segmentation mask $B$ as \begin{equation}
dice=\frac{2|A \cap B|}
 {|A|+|B|},
\label{dice}
\end{equation}
where $\cap$ denotes the intersection (common pixels) of the two masks and $|A|$ denotes the number of pixels in $A$. In classification terms, the union corresponds to TP, whereas $A = TP + FN$ and $B = TP + FP$. Thus, dice coefficient is equivalent to F1 score. 
Dice coefficient was used to evaluate welding defect segmentation algorithms in \cite{W1} and fiber segmentation in \cite{konopczynski2017reference} and \cite{konopczynski2018fully}.

\subsubsection{\ac{IOU} -segmentation}
\label{sssec:iou}
IoU can be used to evaluate segmentation methods by comparing the ground truth segmentation mask $A$ with the predicted mask $B$ as
\begin{equation}
IoU=\frac{|A \cap B|} {|A \cup B|} = \frac{TP}{TP + FP + FN}.
\end{equation} 
Comparing the above equation with Dice Coefficient (F1 score) shows that they are similar. In fact, 
\begin{equation}
dice = \frac{2*IoU}{IoU + 1}.
\end{equation}
Therefore, it is not meaningful to use both IoU and Dice Coefficient at the same time.

IoU was used in \cite{A6} to evaluate additive manufacturing defect segmentation. In \cite{A1}, the same task was performed as 3D segmentation, and also here, IoU was used as the performance metric. In this case, $A$ and $B$ consist of voxels. This metric was used in \cite{abdelkader2021segmentation} with the name \emph{Jaccard Index} to evaluate the segmentation performance in welding defects inspection.

\subsubsection{Clustering Metrics -segmentation}
\label{ssec:clusteringindices}
Segmentation can be also seen as a clustering task, where regions corresponding to different objects should be assigned to different clusters. Therefore, different clustering metrics can be used for evaluating the performance of segmentation methods. In \cite{S2}, the Davies–Bouldin index \cite{davies1979dbindex}, Calinski-Harabasz index \cite{calinski1974chindex}, Dunn index \cite{dunn1974dunnindex} and Hartigan index \cite{hartigan1975clustering} were used to compare X-Ray baggage image segmentation methods. In \cite{M4}, Adjusted Rand Index (ARI) was used to evaluate a 3D fiber instance segmentation method by considering voxels as items to be clustered into uniform instances.

\subsection{Detection Metrics}
Object detection methods typically provide as output the bounding boxes for the detected objects, their predicted classes, and the corresponding confidence values for the predicted classes. Each image can contain multiple objects and each object must be first located and then recognized. Therefore, comparing object detection algorithms is more complicated than comparing classification methods.
A commonly adopted approach is to report \ac{MAP} at selected \ac{IOU} thresholds as described below.

\subsubsection{Intersection over Union (IoU) -detection}
\ac{IOU} defined in Section \ref{sssec:iou} is broadly used also for the evaluation of object detection methods, but not as an independent evaluation metric. A threshold in the \ac{IOU} between a detected bounding box and a ground truth bounding box is used to decide whether the detected bounding box is considered to match the ground-truth bounding box. These matches between detected and ground-truth bounding boxes are then used in \ac{MAP} computation.

\subsubsection{\acf{MAP} -detection} 
In object detection, \ac{MAP} is computed from class-specific precision-recall curves as in classification (see Section \ref{ssec:map}), but the difference is in computing the ranked lists used for computing the precision and recall values. In detection, an \ac{IOU} threshold is first selected (a value of 0.5 is commonly used). Then, the detected bounding boxes for a class are first ranked based on their confidence values. They are then assigned as true positives or false positives by comparing them with the ground-truth bounding boxes for the class and using the selected \ac{IOU} threshold. If there are several detected bounding boxes corresponding to a single ground-truth bounding box, the detected bounding box with the highest \ac{IOU} is considered as true positive, all the others as false positives. The ranked list with the corresponding true/false positive assignments can be used to compute the class-specific \ac{AP} scores and, finally, the \ac{MAP} score as described in Section \ref{ssec:map}.

\subsubsection{Soft-IoU}

In defect detection, the defects such as gas cavities in casting parts cannot be considered independent objects similar to animals or humans, but it can be equally correct to annotate cavities close to each other with a single bounding box or several separate bounding boxes. While \ac{IOU} criterion is used to match a single output bounding box with a single ground-truth bounding box at a time, Soft-IoU algorithm proposed in \cite{C3} can match multiple bounding boxes with a single bounding box or vice versa. \ac{MAP} can be computed using Soft-IoU when it is more suitable than \ac{IOU}.

\subsubsection{Object Localization Accuracy}
\label{ssec:localizationaccuracy}
Object localization accuracy is used for evaluating object localization heatmaps using ground-truth bounding boxes. If the pixel of a maximum response is inside one of the bounding boxes for the specific class, the detection is considered true positive. Otherwise, it is considered false positive. Finally, the object localization accuracy is computed as $\frac{TP}{TP+FP}$. This metric is used in a class-specific object localization task for SIXray dataset \cite{sixray} described in Section~\ref{ssec:sixray}

\subsection{Speed Metrics}
There are several time-based metrics to assess the speed of the models, such as \emph{Frames Per Second (FPS)} that was used in \cite{C5} to evaluate the speed of the model. The other common speed metrics are \emph{Training time} \cite{A1, S3, S6} and \emph{Evaluation time} \cite{C9, C8, S3, S6}, which can be calculated based on \ac{CPU} and \ac{GPU}-based executions.

\section{Datasets}
\label{sec:datasets}

\begin{table*}[htbp]
\caption{\label{Dataset_table} X-ray datasets in industrial and security applications }
\footnotesize
\centering
\resizebox{\textwidth}{!}{
\begin{tabular}{lllllp{2.9cm}p{4cm}ll}

\cline{1-5}
\rowcolor[HTML]{9B9B9B} 
\multicolumn{2}{l}{\cellcolor[HTML]{9B9B9B}}                          & \multicolumn{3}{l}{\cellcolor[HTML]{9B9B9B}Number of images} & \cellcolor[HTML]{9B9B9B}                                  & \cellcolor[HTML]{9B9B9B}                                                                          & \multicolumn{2}{l}{\cellcolor[HTML]{9B9B9B}}                                                                          \\ \cline{3-5}
\rowcolor[HTML]{9B9B9B} 
\multicolumn{2}{l}{\multirow{-2}{*}{\cellcolor[HTML]{9B9B9B}Dataset}} & Train             &  \vtop{\hbox{\strut Test}\hbox{\strut (validation)}}              & Total                & \multirow{-2}{*}{\cellcolor[HTML]{9B9B9B}Resolution}      & \multirow{-2}{*}{\cellcolor[HTML]{9B9B9B}Applications}                                            & \multicolumn{2}{l}{\multirow{-2}{*}{\cellcolor[HTML]{9B9B9B}Highlights}}                                              \\ \hline
\rowcolor[HTML]{FFFFFF} 
\multicolumn{2}{l}{\cellcolor[HTML]{FFFFFF}CoCr AM XCT}               & -                 & -                 & 4,350                & 1012 × 1012                                               & -                                                                                  & \multicolumn{2}{l}{\cellcolor[HTML]{FFFFFF}Gray-scale X-ray, 2D and 3D}                                                   \\ \hline
\rowcolor[HTML]{C0C0C0} 
\cellcolor[HTML]{C0C0C0}                             & Casting        & -                 & -                 & 2,727                & 256 × 256 to 768 × 572                                    & Defect detection, Geometry estimation,   Defect simulation, Image restoration                     & \multicolumn{2}{l}{\cellcolor[HTML]{C0C0C0}}                                                                          \\ \cline{2-7}
\rowcolor[HTML]{C0C0C0} 
\cellcolor[HTML]{C0C0C0}                             & Welding        & -                 & -                 & 88                   & Highly variable                                           & Defect detection, Defect simulation                                                               & \multicolumn{2}{l}{\cellcolor[HTML]{C0C0C0}}                                                                          \\  \cline{2-7}
\rowcolor[HTML]{C0C0C0} 
\multirow{-5}{*}{\cellcolor[HTML]{C0C0C0}GDXray}     & Baggage        & -                 & -                 & 8,150                & Highly variable                                           & Object detection, Object detection in sequential   views, Object classification                 & \multicolumn{2}{l}{\multirow{-5}{*}{\cellcolor[HTML]{C0C0C0} \vtop{\hbox{\strut Gray-scale X-ray,}\hbox{\strut For research and educational  }\hbox{\strut purposes only}} }} \\ \hline
\rowcolor[HTML]{FFFFFF} 
\cellcolor[HTML]{FFFFFF}                             & SIXray10       & 80 \%             & 20 \%             & 98,219               & \cellcolor[HTML]{FFFFFF}                                  & \cellcolor[HTML]{FFFFFF}                                                                          & \multicolumn{2}{l}{\cellcolor[HTML]{FFFFFF}}                                                                          \\ \cline{2-5}
\rowcolor[HTML]{FFFFFF} 
\cellcolor[HTML]{FFFFFF}                             & SIXray100      & 80 \%             & 20 \%             & 901,829              & \cellcolor[HTML]{FFFFFF}                                  & \cellcolor[HTML]{FFFFFF}                                                                          & \multicolumn{2}{l}{\cellcolor[HTML]{FFFFFF}}                                                                          \\ \cline{2-5}
\rowcolor[HTML]{FFFFFF} 
\cellcolor[HTML]{FFFFFF}                             & SIXray1000     & 80 \%             & 20 \%             & 1,051,302            & \cellcolor[HTML]{FFFFFF}                                  & \cellcolor[HTML]{FFFFFF}                                                                          & \multicolumn{2}{l}{\cellcolor[HTML]{FFFFFF}}                                                                          \\ \cline{2-5}
\rowcolor[HTML]{FFFFFF} 
\multirow{-4}{*}{\cellcolor[HTML]{FFFFFF}SIXray}     & total          & 80 \%             & 20 \%             & 1,059,231            & \multirow{-4}{*}{\cellcolor[HTML]{FFFFFF}Highly variable} & \multirow{-4}{*}{\cellcolor[HTML]{FFFFFF} \vtop{\hbox{\strut Image-level classification,}\hbox{\strut Object-level localization}}} & \multicolumn{2}{l}{\multirow{-4}{*}{\cellcolor[HTML]{FFFFFF}Pseudo-colored X-ray}}                                             \\ \hline
\rowcolor[HTML]{C0C0C0} 
\multicolumn{2}{l}{\cellcolor[HTML]{C0C0C0}OPIXray}                   & 7,109             & 1,776             & 8,885                & 1225 × 954                                                & Object detection, Object occlusion                                                                & \multicolumn{2}{l}{\cellcolor[HTML]{C0C0C0} \vtop{\hbox{\strut Pseudo-colored X-ray,}\hbox{\strut Only for academic purposes}} }                                   \\ \hline
\rowcolor[HTML]{FFFFFF} 
\multicolumn{2}{l}{\cellcolor[HTML]{FFFFFF}PIDray}                   & 29,457             & 18,220             & 47,677                & Variable                                                & Classification, Object Detection, Instance Segmentation                                                                & \multicolumn{2}{l}{\cellcolor[HTML]{FFFFFF} \vtop{\hbox{\strut Pseudo-colored X-ray,}\hbox{\strut Only for academic purposes}} }                                   \\ \hline
\rowcolor[HTML]{C0C0C0} 
\multicolumn{2}{l}{\cellcolor[HTML]{C0C0C0}HiXray}                   & \vtop{\hbox{\strut82,452}\hbox{\strut items}}             & \vtop{\hbox{\strut20,476}\hbox{\strut items}}             & \vtop{\hbox{\strut102,928}\hbox{\strut items}}                & \vtop{\hbox{\strut 1200 × 900 (average),}\hbox{\strut 1200 × 1040 (maximum)}}                                                  & Object detection (small object and occluded object detection)                                                               & \multicolumn{2}{l}{\cellcolor[HTML]{C0C0C0} \vtop{\hbox{\strut Pseudo-colored X-ray,}\hbox{\strut Only for academic purposes}} }                                   \\ \hline
\rowcolor[HTML]{FFFFFF} 
\multicolumn{2}{l}{\cellcolor[HTML]{FFFFFF}CLCXray}                   & 80 \%             & \vtop{\hbox{\strut 10 \% }\hbox{\strut (10 \%)}}           &  9,565                & \vtop{\hbox{\strut Between 373 × 200}\hbox{\strut and 732 × 1280}}                                                & Object Detection                     & \multicolumn{2}{l}{\cellcolor[HTML]{FFFFFF} \vtop{\hbox{\strut Pseudo-colored X-ray,}\hbox{\strut Only for academic purposes}} }                                   \\ \hline
\end{tabular}}
\end{table*}

Datasets have a central role in the development of CV methodologies. They are necessary not only to adjust or train the models, but the availability of a public dataset also makes it possible to have a fair comparison among the performance of different CV methodologies. As seen in our review of CV studies on industrial or security-related X-ray images, most of the studies in the field used lab and industrial environment datasets that are not publicly available. In fact, in our opinion, this is one of the main reasons that research on this topic is not advancing at the same pace as other topics in CV where a variety of publicly available datasets exists widely used. However, there are a few publicly available datasets that can be used to develop new methodologies and compare their performance with previously proposed approaches. The main characteristics of these datasets are summarized in Table \ref{Dataset_table} and more details for each dataset are provided in the following.

\subsection{CoCr AM XCT}
\label{ssec:amdataset}
The CoCr AM XCT dataset is introduced in \cite{D1-CoCr-Introduce} and it is available on \cite{D1-CoCr-Source}. The dataset consists of 4,350 images of five cylindrical additive manufacturing specimens. In each specimen, a different minor variation along with geometric magnifications is applied leading to small variations in voxel sizes. A 4-times optical magnification is used on all the specimens. Furthermore, a different exposure time is adopted for each sample. 

\begin{figure}[]
\centering
\includegraphics[scale=.25]{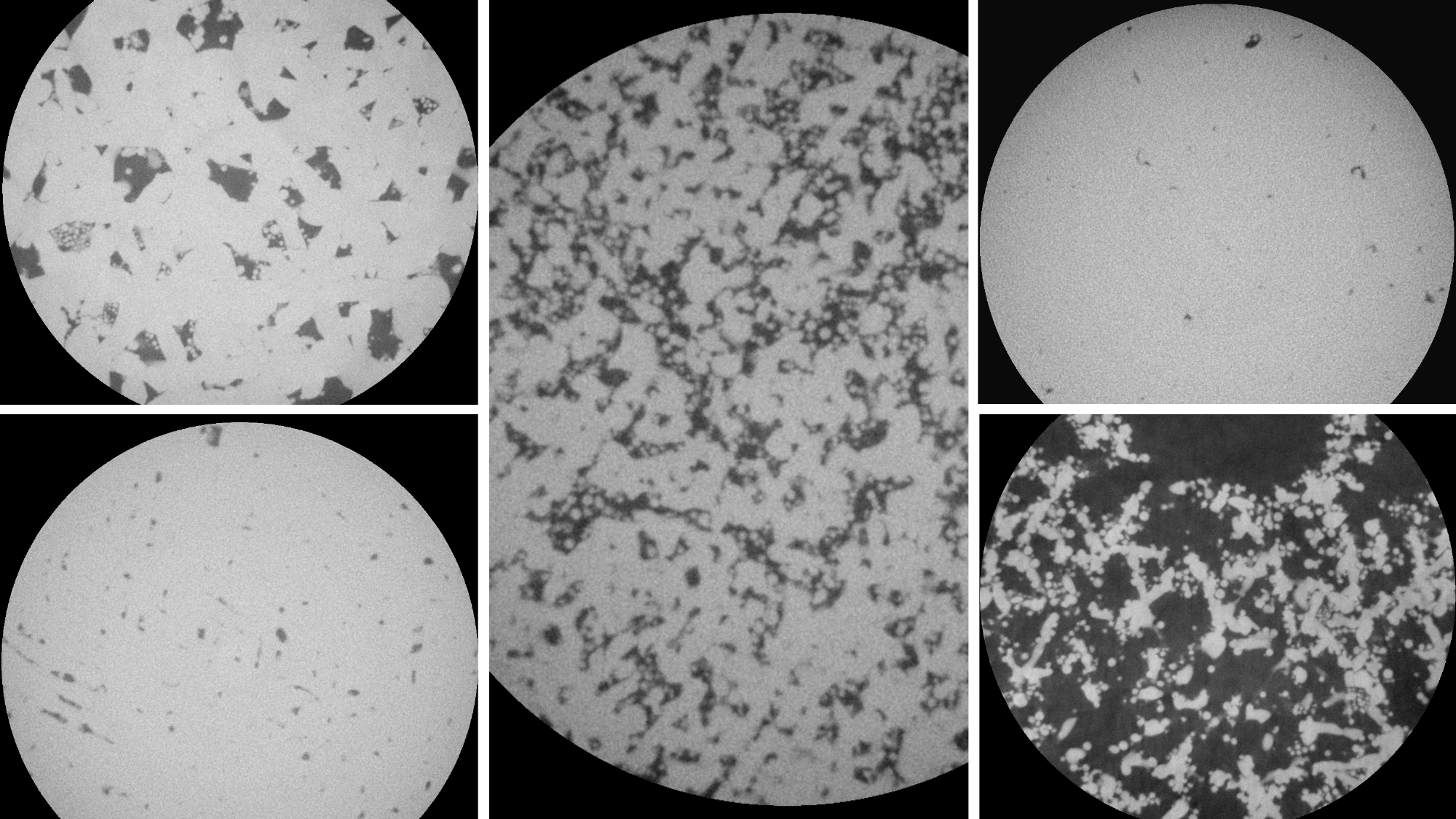}
\caption{Samples of X-ray images from cylindrical AM specimens, available in CoCr AM XCT dataset \cite{D1-CoCr-Introduce}.}
\label{CoCr-img}
\end{figure}

Sample images from CoCr AM XCT dataset are shown in Fig. \ref{CoCr-img}, including images from all five specimens. It should be noted that this dataset was not provided for CV purposes and there is no ground-truth labeling. In \cite{D1-CoCr-Introduce}, the Bernsen local thresholding method \cite{bernsen1986dynamic} was used for defect segmentation and this segmentation was later used as the ground-truth segmentation mask in \cite{A1} for evaluating the performance a fully-convolutional 3D segmentation network. However, as convolutional neural networks have the potential to outperform simple thresholding-based methods such as the one used for the ground-truth generation, it cannot be guaranteed that this evaluation protocol leads to a fair comparison between more advanced methods.

\subsection{GDXray+}
The GDXray+ dataset \cite{gdxray} provides a collection of more than 21,100 X-ray images to develop, test, and evaluate CV and image analysis methods. The dataset is named GDXray according to the name of the Machine Intelligence Group performing the data collection (GRIMA X-ray database). The data can be used freely only for research and education purposes.

\begin{figure}[!t]
\centering
\includegraphics[width=\columnwidth]{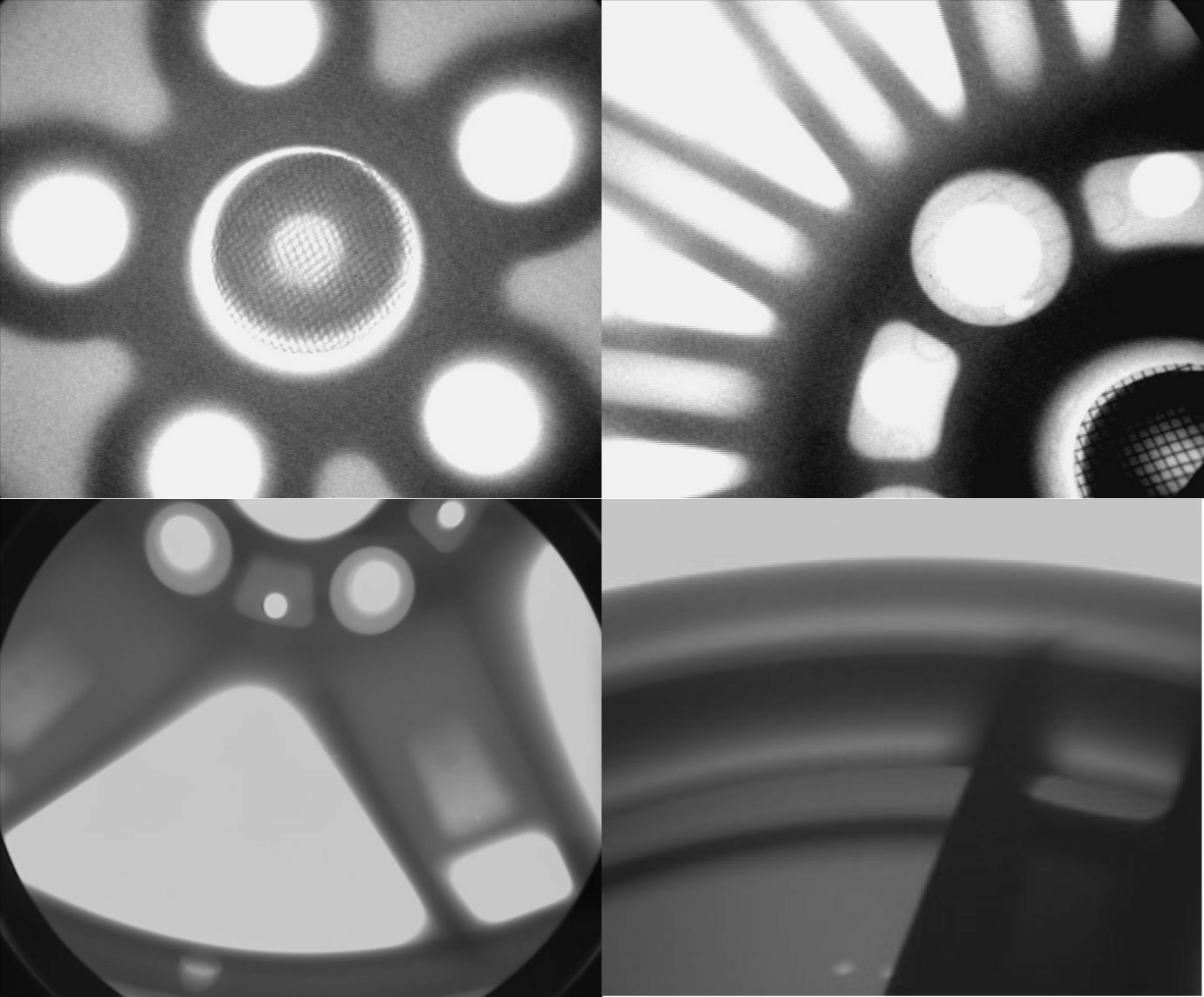}
\caption{Samples of X-ray images from casting specimens, available in GDXray data \cite{gdxray}.}
\label{GDX-ca-img}
\end{figure}

\begin{figure}[!t]
\centering
\includegraphics[width=\columnwidth]{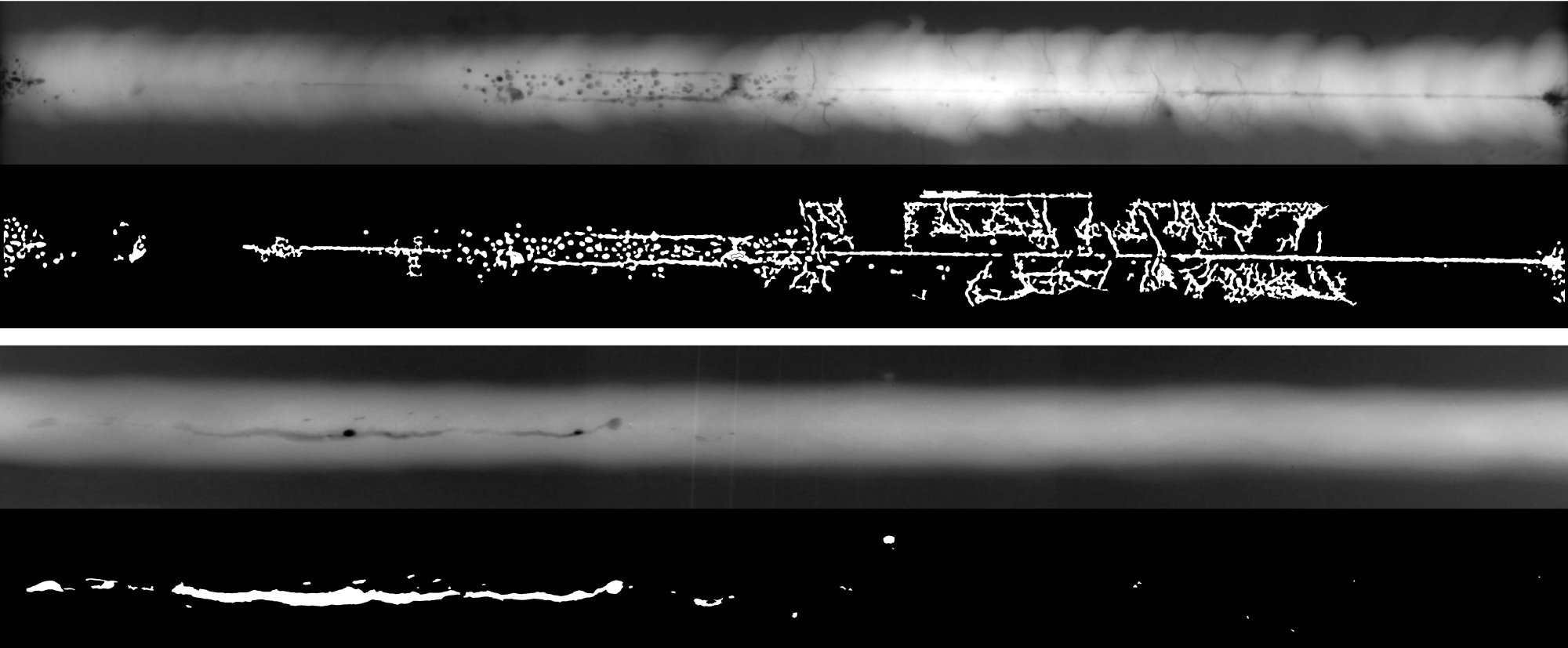}
\caption{Samples of X-ray images from welding specimens and their segmentation, available in GDXray data \cite{gdxray}.}
\label{GDX-we-img}
\end{figure}

\begin{figure}[!t]
\centering
\includegraphics[width=\columnwidth]{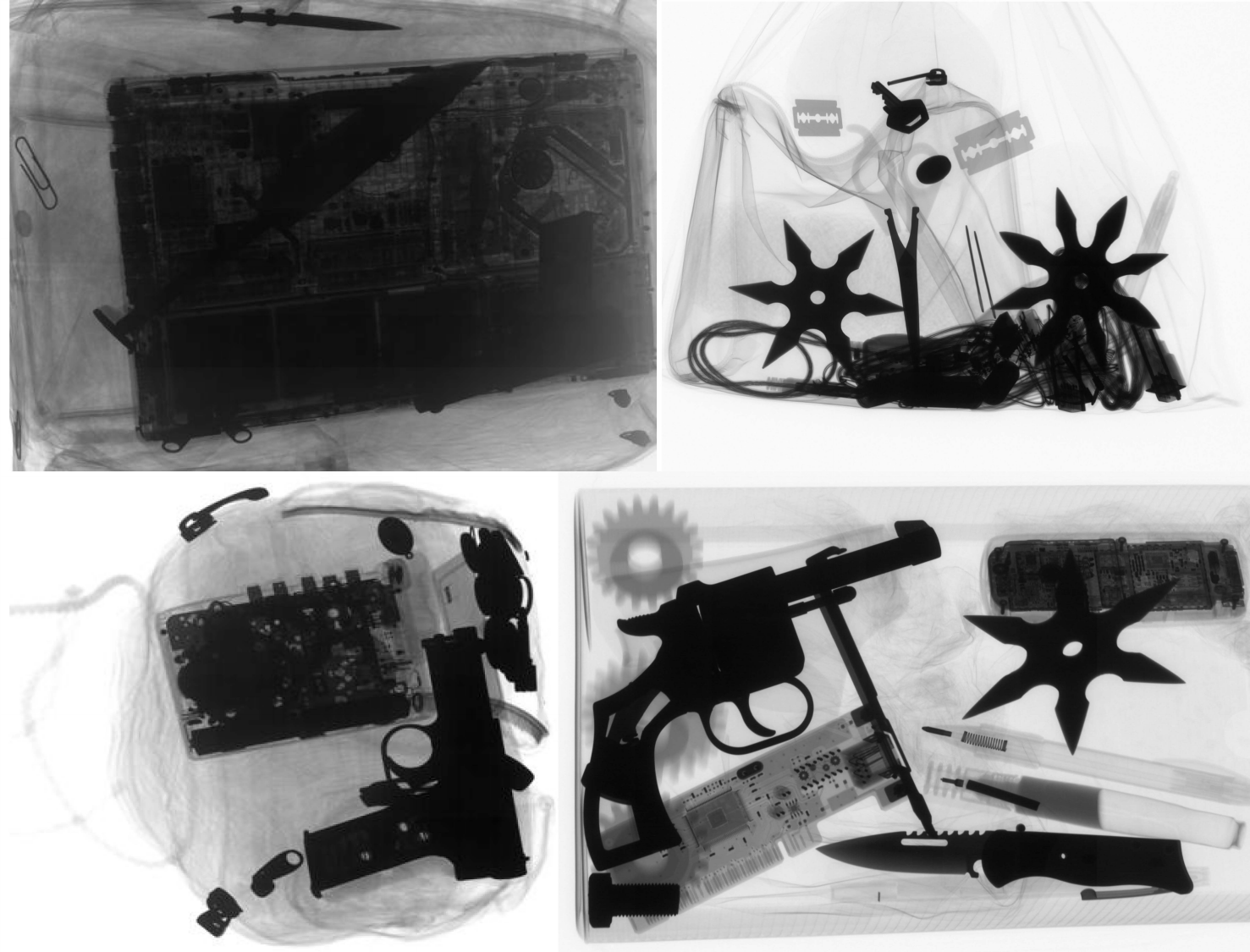}
\caption{Samples of X-ray baggage scans, available in GDXray data \cite{gdxray}.}
\label{GDX-se-img}
\end{figure}

GDXray+ includes five groups of images (casting, welding, security, nature, and setting). Three of these five groups can be considered relevant for our review and they are briefly described in the following:
\begin{itemize}
    \item The \emph{GDXray Casting dataset} contains 2,727 X-ray images mainly from automotive parts, including aluminum wheels and knuckles, many of which contain casting defects. The casting defects in each image are labeled with tight-fitting bounding boxes. The size of the images in the dataset ranges from 256 × 256 pixels to 768 × 572 pixels. This group of X-ray images is arranged in 67 series. The description and applications of each series are available in \cite{gdxray}. Fig. \ref{GDX-ca-img} shows a random collection of the images from the GDXray Casting dataset. While the dataset does not define a default evaluation metric, \ac{MAP} at \ac{IOU} 0.5 has been used in \cite{C8,C9}. A random split into a 80\% training set and a 20\% test set was proposed and made publicly available in \cite{C9}.

    A new dataset obtained from the GDXray Casting data by cropping 32 × 32 pixels patches is introduced in \cite{O3}. This dataset consists of 47,520 X-ray casting images along with their labels.

    \item The \emph{GDXray Welding dataset} includes 88 images of metal pipes welding with porosity defects and it contains pixel-wise ground truth segmentation information for some of the images. This group of X-ray images is arranged in 3 series. The description and applications of each series are available in \cite{gdxray}. Two samples from this dataset along with their ground-truth segmentation are shown in Fig. \ref{GDX-we-img}.

    \item The \emph{GDXray Baggage dataset} contains 8,150 X-ray baggage scans containing both occluded and non-occluded items with marked ground truths for handguns, razor blades, shurikens, and knives. This group of X-ray images is arranged in 77 series. The description and applications of each series are available in \cite{gdxray}. Some image samples from this dataset are shown in Fig. \ref{GDX-se-img}.

\end{itemize}

\subsection{SIXray}
\label{ssec:sixray}
\begin{figure}[!t]
\centering
\includegraphics[scale=.4]{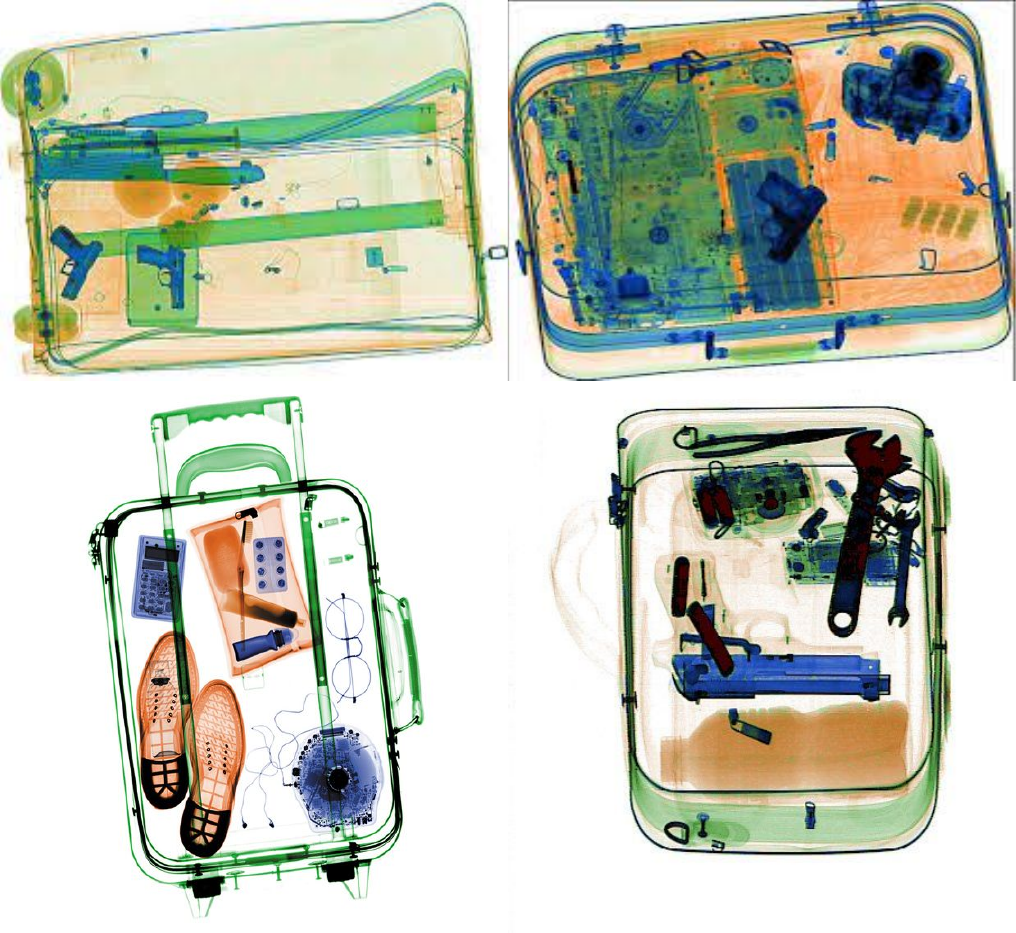}
\caption{Samples of X-ray security images, available in SIXray data \cite{sixray}.}
\label{SIX-img}
\end{figure}

SIXray is a pseudo-color X-ray security inspection dataset introduced in \cite{sixray}. It includes over a million X-ray images that were collected at several subway stations using color-X-ray scanners that assign various colors to different materials. The data is categorized into six common threat groups (gun, wrench, knife, scissors, pliers, and hammer). Some samples from this dataset are shown in Fig. \ref{SIX-img}.
 
To study the impact of class imbalance, three different subsets known as SIXray10, SIXray100, and SIXray1000 are defined as follows: SIXray10 has all 8,929 scans with suspicious items and ten times scans without suspicious items; SIXray100 contains all scans including suspicious items and 100 times non-suspicious scans; SIXray1000 has only 1000 images with suspicious items and all images without suspicious items. Each subset is randomly divided into a training set containing 80$\%$ of the images and a test set containing the remaining 20$\%$ of the images. Image-level annotations provided by human security inspectors are available for the whole dataset, while bounding box annotations of prohibited items are available only for the test datasets.  

The original dataset paper provides results for two different tasks: image-level classification and object localization. Both tasks are evaluated separately for each class. For image-level classification, the methods should rank the test images based on their probability to contain a specific object and the results are evaluated using mAP (see Section \ref{ssec:map}) similar to Pascal VOC classification challenge \cite{everingham2010pascal}. For object localization, the evaluated methods produce heatmaps for each class separately and the performance is evaluated using object localization accuracy (see Section \ref{ssec:localizationaccuracy} as in \cite{zhang2018top}).

\subsection{OPIXray}
OPIXray \cite{opixray} is a pseudo-color X-ray security dataset of occluded prohibited items. The backgrounds of all samples are scanned
by the security inspection machine and the prohibited items are synthesized into these backgrounds using professional software. The dataset consists of 8,885 X-ray images categorized based on 5 prohibited items from five categories: Straight Knife, Folding Knife, Utility Knife, Multi-tool Knife, and Scissors. All prohibited items are annotated manually with a bounding box by a professional inspector from an
international airport. Each image contains at least one prohibited item, while some have more, and images are stored in JPG format with a resolution of 1225 × 954.
Some samples from this dataset are shown in Fig.~\ref{OPIX-img}.

The dataset is partitioned into a
80\% training set and a 20\% test set. Furthermore, the test set is divided into three subsets (OL1-3) containing prohibited items with different occlusion levels. The task to be performed is object detection and the evaluation metric used in \cite{opixray} is \ac{MAP} with 0.5 \ac{IOU} threshold.

\subsection{PIDray}
\begin{figure}[!t]
\centering
\includegraphics[width=\columnwidth]{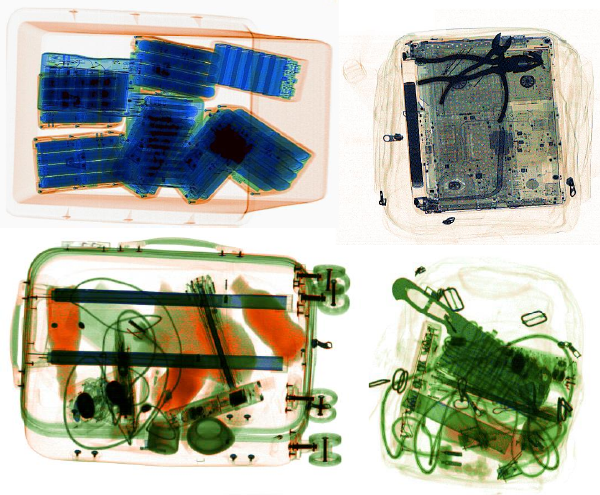}
\caption{Samples of X-ray security images, available in PIDray data \cite{PIDray}.}
\label{PIDray-img}
\end{figure}

PIDray \cite{PIDray} is a large X-ray dataset including 47,677 real security images, each of which contains at least one prohibited item. Some prohibited items have been deliberately hidden. The images are collected from different scenarios including airports, railway stations, and subway stations using three different security inspection machines from different manufacturers resulting in a variety of sizes, colors, and resolutions. The prohibited item categories are knife, gun, scissors, lighter, sprayer, baton, wrench, pliers, hammer, handcuffs, power bank, and bullet. Some samples of this dataset are shown in Fig. \ref{PIDray-img}. 

The dataset is divided into a 60\% train set and a 40\% test set. In addition, the data is split into three groups, namely easy, hard, and hidden. The images are annotated with both bounding boxes and segmentation masks. Therefore, this dataset can be used for classification, object detection, and instance segmentation. The performance metrics used in \cite{PIDray} are Average Precision and Average Recall averaged over 12 classes and 10 \ac{IOU} thresholds between 0.5 and 0.95.

\subsection{HiXray}
\begin{figure}[!t]
\centering
\includegraphics[width=\columnwidth]{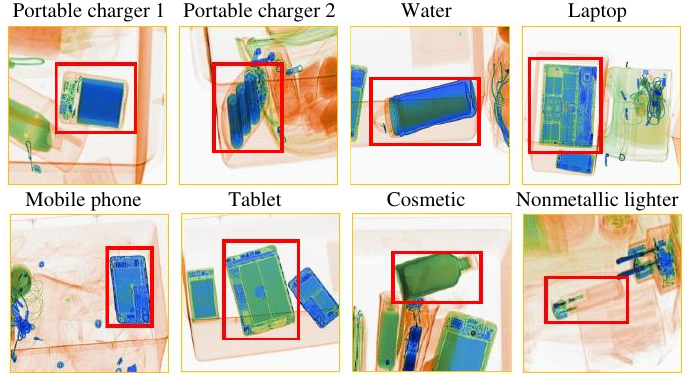}
\caption{Samples of X-ray security images, available in HiXray data \cite{HiXray}.}
\label{HiXray-img}
\end{figure}

The High-quality X-ray (HiXray) security inspection image dataset was introduced in \cite{HiXray}. It is the largest high-quality dataset for prohibited item detection, and it contains 45,364 pseudo-color X-ray images with 102,925 common prohibited items which are categorized into 8 classes, namely portable charger 1 (lithium-ion prismatic cell), portable charger 2 (lithium-ion cylindrical cell), water, laptop, mobile phone, tablet, cosmetic, and nonmetallic lighter. The images are collected from a real-world airport security inspection and bounding box annotations are provided manually by professional security inspectors. The images are in JPG format with an average resolution of 1200 × 900, and on average, each image has 2.27 prohibited items. A sample of each class in HiXray dataset is shown in Fig. \ref{HiXray-img}.

The dataset is divided into training and test subsets with a 4:1 ratio. The dataset is proposed for detection tasks and can be used for more specific tasks such as small object or occluded object detection. The evaluation metric used in \cite{HiXray} is \ac{MAP} with 0.5 \ac{IOU} threshold.

\subsection{CLCXray}
\label{ssec:clcxray}
The Cutters and Liquid Containers X-ray Dataset (CLCXray) \cite{CLCXray} focuses particularly on the overlap problem in security images. While OPIXray dataset also focuses on the overlap problem, CLCXray has more overlap
between objects and similar backgrounds, as well as
overlap between multiple objects. In addition, the images in OPIXray
are synthetic, while CLCXray contains real images. Overall, compared to other security datasets, CLCXray
has the most labeled images, the most labeled threat objects, the most threat
categories, and more accurate bounding box annotations. 

\begin{figure*}[!t]
\centering
\includegraphics[width=0.8\textwidth]{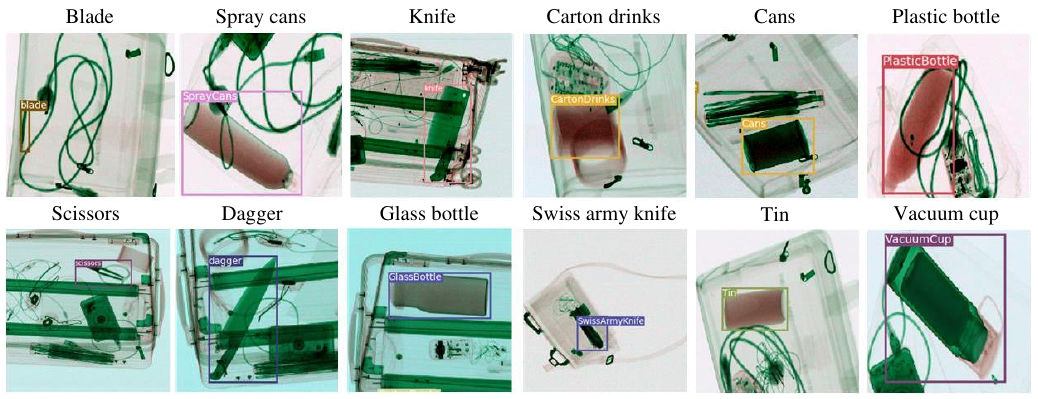}
\caption{Samples of X-ray security images, available in CLCXray data \cite{CLCXray}.}
\label{CLCXray-img}
\end{figure*}

There are 9,565 pseudo-color X-ray security images that consist of 4,543 images collected from real subway scenes and 5,022 simulated images from manually designed baggages. The images are labeled by professionals in 12 categories including 5 classes of cutters (blade, knife, dagger, scissors, Swiss army knife), and 7 classes of liquid containers (cans, carton drinks, plastic bottle, glass bottle, vacuum cup, tin, spray cans). In total, there are more than 20,000 potentially dangerous items in the dataset resulting in an average of more than two items per image. The images have resolutions between 373 × 200 and 732 × 1280 pixels. A sample of each category in CLCXray dataset is shown in Fig. \ref{CLCXray-img}. 

The dataset is divided using an 8:1:1 ratio into training, validation, and testing sets. Annotations are provided in COCO format. For evaluation, CLCXray adopts the COCO evaluation metrics \cite{coco}:  
$mAP$ is the mean average precision computed across 10 \ac{IOU} levels of 0.5:0.05:0.95, $mAP_{50}$ is computed at a single \ac{IOU} of
0.5. $mAP_{75}$ is computed at a single \ac{IOU} of 0.75, $mAP_s$ in the \ac{MAP} for small objects (area < 322), $mAP_m$ is
the \ac{MAP} for medium objects (322 < area < 962), and $mAP_l$ is the \ac{MAP} for large objects (962 < area).

\begin{table*}[bp!]
\caption{\vtop{\hbox{\strut Performance comparison on CoCr AM XCT dataset}\hbox{\strut \normalfont{$mIoU$: mean Intersection over Union}}}} \label{CoCr}
\footnotesize
\centering
\begin{tabular}{cccc}
\hline
\rowcolor[HTML]{AEAAAA} 
\cellcolor[HTML]{AEAAAA}                                              & \cellcolor[HTML]{AEAAAA}                         & Evaluation metric & Time metric                   \\ \cline{3-4} 
\rowcolor[HTML]{AEAAAA} 
\multirow{-2}{*}{\cellcolor[HTML]{AEAAAA}Reference}                   & \multirow{-2}{*}{\cellcolor[HTML]{AEAAAA}Method} & mIoU            & Training   Time (hour) on GPU \\ \hline
\rowcolor[HTML]{FFFFFF} 
\cellcolor[HTML]{FFFFFF}                                              & 3D   U-Net with Conv+BN+ReLU                     & 0.863           & \textbf{6.58}                          \\  \cline{2-4}
\rowcolor[HTML]{FFFFFF} 
\cellcolor[HTML]{FFFFFF}                                              & 3D   U-Net with Conv+ReLU+GN                     & 0.881           & 14.00                         \\ \cline{2-4}
\rowcolor[HTML]{FFFFFF} 
\multirow{-3}{*}{\cellcolor[HTML]{FFFFFF}\cite{A1}} & \textbf{Residual   Symmetric 3D U-Net}                    & \textbf{0.884}           & 19.97                         \\ \hline
\end{tabular}
\end{table*}

\section{Performance Comparison}
\label{sec:comparison}

Due to the lack of available public datasets for industrial X-ray image processing, most studies have used private data, which makes it difficult to compare and verify the performance of different approaches for different tasks. Furthermore, even studies using public datasets use different evaluation protocols and metrics. Several works have not reproduced the results of some of the methods used in the comparisons but simply transferred them from prior work. Thus, direct comparison with the exact same experimental protocol is not possible. In some cases, this has even led to directly comparing results for different tasks (e.g., classification and detection). 

In this section, we have collected results reported in previous works using public datasets. These results are divided based on the datasets and presented in Tables \ref{CoCr} to \ref{CLCXray}. Due to the above-mentioned problems, we report the results collected from different studies in separate blocks of rows with different background colors. Furthermore, we indicate where the comparative results have been collected from different studies, and in our discussion, we point out some of the clearest problems we observed.

\subsection{CoCr AM XCT}
As can be seen in Table \ref{CoCr}, only one paper employed this dataset \cite{A1}. The paper assessed three variants of 3D U-Net on the dataset. The first one is a 3D U-Net with convolutional, batch normalization (BN), and rectified linear unit (ReLU) layers. The second one uses convolutional, group normalization (GN), and ReLU layers, and the last one is a residual symmetric 3D-Net. mIoU and training time on GPU were chosen as the accuracy and time metrics and the results show that the residual symmetric 3D U-Net achieved the best accuracy, but it was slower to train. As the training time is not connected to the real-time operation, the computational time can be neglected, and the higher accuracy would be the selection index among these three methods. However, as mentioned in \ref{ssec:amdataset}, the ground-truth annotations were generated using a simple thresholding approach and it is unclear whether this leads to a meaningful comparison of the more advanced methods. 

\subsection{GDXray Casting}
As can be seen in Table \ref{GDXray-c}, two papers employed this dataset \cite{C9, C8}. In the first one \cite{C9}, the performances of six different methods including Sliding window, Faster R-CNN VGG-16, Faster R-CNN ResNet-101, R-FCN ResNet-101, SSD VGG-16, and SSD ResNet-101 were compared. Faster R-CNN with ResNet-101 as the backbone achieved the highest accuracy by 0.921 mAP. The best evaluation time on both CPU and GPU was obtained by SSD architecture with VGG-16 backbone. The second study from the same authors \cite{C8} proposed using a Mask R-CNN-based approach that produces both a segmentation mask and bounding boxes. Comparative results with different learning strategies 
(random weight initialization, pretrained ImageNet weights, and ImageNet weights with pretraining on MS-COCO dataset) are shown in Table \ref{GDXray-c}. Here, \ac{MAP} is reported for both bounding boxes and mask predictions, while the ground truth annotations consist of bounding boxes.

\begin{table*}[htbp]
\caption{\label{GDXray-c} \vtop{\hbox{\strut Performance comparison on GDXray Casting dataset}\hbox{\strut \normalfont{$mAP_{bbox}$: mean Average Precision at 0.5 \ac{IOU} of for bounding box predictions}}
\hbox{\strut \normalfont{$mAP_{mask}$: mean Average Precision at 0.5 \ac{IOU} of for mask predictions}}}}
\footnotesize
\centering
\begin{tabular}{ccccccc}
\hline
\rowcolor[HTML]{AEAAAA} 
\cellcolor[HTML]{AEAAAA}                                              & \multicolumn{2}{c}{\cellcolor[HTML]{AEAAAA}}                                                                                                & \multicolumn{2}{c}{\cellcolor[HTML]{AEAAAA}Evaluation metric}                                           & \multicolumn{2}{c}{\cellcolor[HTML]{AEAAAA}Time metric}              \\ \cline{4-7} 
\rowcolor[HTML]{AEAAAA} 
\cellcolor[HTML]{AEAAAA}                                              & \multicolumn{2}{c}{\cellcolor[HTML]{AEAAAA}}                                                                                                & \cellcolor[HTML]{AEAAAA}                          & \cellcolor[HTML]{AEAAAA}                          & \multicolumn{2}{c}{\cellcolor[HTML]{AEAAAA}Evaluation   Time (s) on} \\ \cline{6-7} 
\rowcolor[HTML]{AEAAAA} 
\multirow{-3}{*}{\cellcolor[HTML]{AEAAAA}Reference}                   & \multicolumn{2}{c}{\multirow{-3}{*}{\cellcolor[HTML]{AEAAAA}Method}}                                                                        & \multirow{-2}{*}{\cellcolor[HTML]{AEAAAA}$mAP_{bbox}$} & \multirow{-2}{*}{\cellcolor[HTML]{AEAAAA}$mAP_{mask}$} & GPU                               & CPU                              \\ \cline{1-1} \cline{4-7} 
\rowcolor[HTML]{FFFFFF} 
\cellcolor[HTML]{FFFFFF}                                              & \multicolumn{2}{c}{\cellcolor[HTML]{FFFFFF}Sliding window}                                                                                  & 0.461                                             & -                                                 & 0.231                             & 2.231                            \\ \cline{2-7}
\rowcolor[HTML]{FFFFFF} 
\cellcolor[HTML]{FFFFFF}                                              & \multicolumn{2}{c}{\cellcolor[HTML]{FFFFFF}Faster R-CNN VGG-16}                                                                             & 0.865                                             & -                                                 & 0.438                             & 7.291                            \\ \cline{2-7}
\rowcolor[HTML]{FFFFFF} 
\cellcolor[HTML]{FFFFFF}                                              & \multicolumn{2}{c}{\cellcolor[HTML]{FFFFFF}Faster R-CNN ResNet-101}                                                                         & 0.921                                             & -                                                 & 0.512                             & 9.319                            \\ \cline{2-7}
\rowcolor[HTML]{FFFFFF} 
\cellcolor[HTML]{FFFFFF}                                              & \multicolumn{2}{c}{\cellcolor[HTML]{FFFFFF}R-FCN ResNet-101}                                                                                & 0.875                                             & -                                                 & 0.375                             & 3.721                            \\ \cline{2-7}
\rowcolor[HTML]{FFFFFF} 
\cellcolor[HTML]{FFFFFF}                                              & \multicolumn{2}{c}{\cellcolor[HTML]{FFFFFF}SSD VGG-16}                                                                                      & 0.697                                             & -                                                 & \textbf{0.025}                             & \textbf{0.088}                            \\ \cline{2-7}
\rowcolor[HTML]{FFFFFF} 
\multirow{-6}{*}{\cellcolor[HTML]{FFFFFF}\cite{C9}} & \multicolumn{2}{c}{\cellcolor[HTML]{FFFFFF}SSD ResNet-101}                                                                                  & 0.762                                             & -                                                 & 0.051                             & 0.141                            \\ \hline
\rowcolor[HTML]{E7E6E6} 
\cellcolor[HTML]{E7E6E6}                                              & \cellcolor[HTML]{E7E6E6}                                                   & Xavier Initialization (Random)                                 & 0.651                                             & 0.420                                             & -                                 & -                                \\ \cline{3-7}
\rowcolor[HTML]{E7E6E6} 
\cellcolor[HTML]{E7E6E6}                                              & \cellcolor[HTML]{E7E6E6}                                                   & Pretrained ImageNet Weights                                    & 0.874                                             & 0.721                                             & -                                 & -                                \\ \cline{3-7}
\rowcolor[HTML]{E7E6E6} 
\multirow{-3}{*}{\cellcolor[HTML]{E7E6E6}\cite{C8}} & \multirow{-3}{*}{\cellcolor[HTML]{E7E6E6}} ResNet-101 & \vtop{\hbox{\strut Pretrained ImageNet Weights}\hbox{\strut (Pretraining on MS-COCO dataset)}}   & \textbf{0.957}                                             & \textbf{0.930}                                             & 0.165                             & 6.240                            \\ \hline
\end{tabular}
\end{table*}

\begin{table*}[h]
\caption{\label{GDXray-w}\vtop{\hbox{\strut Performance comparison on GDXray Welding dataset}\hbox{\strut \normalfont{$\eta$: Accuracy, $AUC$: Area Under Curve}}}}
\footnotesize
\centering
\begin{tabular}{cccccccccl}
\hline
\rowcolor[HTML]{AEAAAA} 
\cellcolor[HTML]{AEAAAA} & \cellcolor[HTML]{AEAAAA} & \multicolumn{8}{c}{\cellcolor[HTML]{AEAAAA}Evaluation metric} \\ \cline{3-10} 
\rowcolor[HTML]{AEAAAA} 
\cellcolor[HTML]{AEAAAA} & \cellcolor[HTML]{AEAAAA} & \cellcolor[HTML]{AEAAAA} & \cellcolor[HTML]{AEAAAA} & \cellcolor[HTML]{AEAAAA} & \cellcolor[HTML]{AEAAAA} & \cellcolor[HTML]{AEAAAA} & \cellcolor[HTML]{AEAAAA} & \cellcolor[HTML]{AEAAAA} & \cellcolor[HTML]{AEAAAA} \\
\rowcolor[HTML]{AEAAAA} 
\multirow{-3}{*}{\cellcolor[HTML]{AEAAAA}Reference} & \multirow{-3}{*}{\cellcolor[HTML]{AEAAAA}Method} & \multirow{-2}{*}{\cellcolor[HTML]{AEAAAA}Precision} & \multirow{-2}{*}{\cellcolor[HTML]{AEAAAA}Recall} & \multirow{-2}{*}{\cellcolor[HTML]{AEAAAA}Specificity} & \multirow{-2}{*}{\cellcolor[HTML]{AEAAAA}$\eta$} & \multirow{-2}{*}{\cellcolor[HTML]{AEAAAA}AUC} & \multirow{-2}{*}{\cellcolor[HTML]{AEAAAA}Dice} & \multirow{-2}{*}{\cellcolor[HTML]{AEAAAA}F1}  & \multirow{-2}{*}{\cellcolor[HTML]{AEAAAA}Experimental protocol} \\ \hline
\rowcolor[HTML]{FFFFFF} 
\cellcolor[HTML]{FFFFFF} & U-Net & - & 0.864 & 0.998 & 0.998 & 0.800 & 0.782 & -  & Random cropping of training data, \\ 
\rowcolor[HTML]{FFFFFF} 
\cellcolor[HTML]{FFFFFF} & eGAN & - & 0.618 & \textbf{0.999} & 0.997 & 0.759 & 0.708 & - & uniform cropping of test data,\\
\rowcolor[HTML]{FFFFFF} 
\multirow{-3}{*}{\cellcolor[HTML]{FFFFFF}\cite{W1}} & Improved U-net & - & 0.860 & 0.999 & 0.998 & 0.884 & 0.818 & -  & details not given \\ \hline
\rowcolor[HTML]{E7E6E6} 
\cellcolor[HTML]{E7E6E6} & Proposed 2-layer network & 0.881 & 0.881 & - & 0.886 & - & - & 0.880 & 32x32 cropped patches,\\ 
\rowcolor[HTML]{E7E6E6} 
\multirow{-2}{*}{\cellcolor[HTML]{E7E6E6}\cite{W7}} & Proposed 3-layer network & 0.890 & 0.899 & - & 0.898 & - & - & 0.894 & 5-fold cross validation\\ \hline
\end{tabular}
\end{table*}

\subsection{GDXray Welding}
Results for the GDXray Welding dataset are shown in Table \ref{GDXray-w}. An improved U-net was compared in \cite{W1} with U-Net and eGAN, assessing by five accuracy indices (recall/sensitivity, specificity, accuracy, \ac{AUC}, and dice), and on four out of the five indices, it achieved a better performance. Two \ac{DL} networks with two and three hidden layers were proposed in \cite{W7}. The three-hidden layer network achieved a better performance evaluated by Precision, Recall, and F1. As the original dataset contains only 88 large welding images, both of the above studies applied cropping to create a larger dataset. However, different cropping makes the results incomparable. Furthermore, in \cite{C8} transfer learning from casting defect detection to welding defect detection was evaluated using the GDXray Welding dataset with again a different image cropping approach. In this work, a mAP$_{mask}$ of 0.85 was achieved.

The GDXray Welding data was used as a part of the experiments also in \cite{W2}, where a Retina-based network was used for welding defect detection. However, here the dataset used in the experiments contained as also the GDXray Casting data as well as privately collected welding X-ray images, and thus the results are even less comparable to other studies.

\begin{table*}
\caption{\label{GDXray-s}\vtop{\hbox{\strut Performance comparison on GDXray Security Dataset}\hbox{\strut \normalfont{$mAP$: mean Average Precision, $S_p$: Specificity, $\eta$: Accuracy, $AUC$: Area Under Curve, $R$: Recall, $P$: Precision}}}}
\footnotesize
\centering
\resizebox{\textwidth}{!}{
\begin{tabular}{lllllllllp{5cm}}
\hline
\rowcolor[HTML]{AEAAAA} 
\cellcolor[HTML]{AEAAAA} & \cellcolor[HTML]{AEAAAA} & \multicolumn{7}{l}{\cellcolor[HTML]{AEAAAA}Accuracy metric} & \cellcolor[HTML]{AEAAAA} \\
\rowcolor[HTML]{AEAAAA} 
\multirow{-2}{*}{\cellcolor[HTML]{AEAAAA}Reference} & \multirow{-2}{*}{\cellcolor[HTML]{AEAAAA}Method} & $mAP$ & $S_p$ & $\eta$ & $AUC$ & $F1$ & $R$ & $P$ & \multirow{-2}{*}{\cellcolor[HTML]{AEAAAA}Experimental protocols} \\ \hline
\rowcolor[HTML]{FFFFFF} 
\cellcolor[HTML]{FFFFFF} & AISM & - & 0.965 & - & 0.992 & - & 0.985 & - & \cellcolor[HTML]{FFFFFF} \\
\rowcolor[HTML]{FFFFFF} 
\cellcolor[HTML]{FFFFFF} & SURF & - & 0.630 & - & 0.616 & - & 0.656 & - & \cellcolor[HTML]{FFFFFF} \\
\rowcolor[HTML]{FFFFFF} 
\cellcolor[HTML]{FFFFFF} & SIFT & - & 0.830 & - & 0.921 & - & 0.884 & - & \cellcolor[HTML]{FFFFFF} \\
\multirow{-4}{*}{\cellcolor[HTML]{FFFFFF}\cite{S3-15}} & ISM & - & 0.885 & - & 0.955 & - & 0.924 & - & \multirow{-4}{*}{\cellcolor[HTML]{FFFFFF}\begin{tabular}[c]{@{}l@{}}- Training images: 400, testing images: 600\\ - Objects: razor blades, shuriken, handguns\end{tabular}} \\ \hline
\rowcolor[HTML]{E7E6E6} 
\cellcolor[HTML]{E7E6E6} & Faster R-CNN & - & - & 0.984 & - & 0.954 & 0.980 & 0.930 & \cellcolor[HTML]{E7E6E6} \\
\rowcolor[HTML]{E7E6E6} 
\cellcolor[HTML]{E7E6E6} & YOLOv2 & - & - & 0.971 & - & 0.900 & 0.880 & 0.920 & \cellcolor[HTML]{E7E6E6} \\
\rowcolor[HTML]{E7E6E6} 
\multirow{-3}{*}{\cellcolor[HTML]{E7E6E6}\cite{S3-25}} & Tiny YOLO & - & - & 0.890 & - & 0.750 & 0.820 & 0.690 & \multirow{-3}{*}{\cellcolor[HTML]{E7E6E6}\begin{tabular}[c]{@{}l@{}}- 3669 images\\ - Train to test ratio = 8:2\\ - Objects: knife, gun, shuriken, razor blade\end{tabular}} \\ \hline
\rowcolor[HTML]{FFFFFF} 
\cellcolor[HTML]{FFFFFF} & ResNet50+CST & - & 0.989 & 0.968 & 0.987 & 0.918 & 0.886 & 0.953 & \cellcolor[HTML]{FFFFFF} \\
\rowcolor[HTML]{FFFFFF} 
\multirow{-2}{*}{\cellcolor[HTML]{FFFFFF}\begin{tabular}[c]{@{}l@{}}\cite{S3} and\\ \cite{S6}\end{tabular}} & ResNet50+CST & - & 0.965 & 0.983 & 0.993 & 0.984 & 0.997 & 0.971 & \multirow{-2}{*}{\cellcolor[HTML]{FFFFFF}\begin{tabular}[c]{@{}l@{}} - Objects: razor blades, shuriken, handguns, \\ chip, pistol, mobile and knives\end{tabular}} \\ \hline
\rowcolor[HTML]{E7E6E6} 
\cellcolor[HTML]{E7E6E6} & Faster R-CNN & 0.652 & - & - & - & - & - & - & \cellcolor[HTML]{E7E6E6} \\
\rowcolor[HTML]{E7E6E6} 
\multirow{-2}{*}{\cellcolor[HTML]{E7E6E6}\cite{S14}} & Faster R-CNN+PDN & 0.788 & - & - & - & - & - & - & \multirow{-2}{*}{\cellcolor[HTML]{E7E6E6}\begin{tabular}[c]{@{}l@{}}- Data: B0009-B0044 and B0046-B0048\\ - Objects: gun, shuriken, knife\end{tabular}} \\ \hline
\rowcolor[HTML]{FFFFFF} 
\cellcolor[HTML]{FFFFFF} & SSD \,\, \begin{tabular}[r]{@{}l@{}}+random\\ +learned\end{tabular} & \begin{tabular}[c]{@{}l@{}}0.635\\ 0.663\end{tabular} & - & - & - & - & - & - & \cellcolor[HTML]{FFFFFF} \\
\rowcolor[HTML]{FFFFFF} 
\cellcolor[HTML]{FFFFFF} & RefineDet \,\, \begin{tabular}[c]{@{}l@{}}+random\\ +learned\end{tabular} & \begin{tabular}[c]{@{}l@{}}0.709\\ 0.759\end{tabular} & - & - & - & - & - & - & \cellcolor[HTML]{FFFFFF} \\
\rowcolor[HTML]{FFFFFF} 
\cellcolor[HTML]{FFFFFF} & PFPNet \,\, \begin{tabular}[c]{@{}l@{}}+random\\ +learned\end{tabular} & \begin{tabular}[c]{@{}l@{}}0.741\\ 0.771\end{tabular} & - & - & - & - & - & - & \cellcolor[HTML]{FFFFFF} \\
\rowcolor[HTML]{FFFFFF} 
\multirow{-4}{*}{\cellcolor[HTML]{FFFFFF}\cite{W08}} & RFBNet \,\, \begin{tabular}[c]{@{}l@{}}+random\\ +learned\end{tabular} & \begin{tabular}[c]{@{}l@{}}0.568\\ 0.620\end{tabular} & - & - & - & - & - & - & \multirow{-6}{*}{\cellcolor[HTML]{FFFFFF}\begin{tabular}[c]{@{}l@{}}- Data: 200 test images\\ - Objects: knife, gun, shuriken, razor blade\end{tabular}} \\ \hline
\rowcolor[HTML]{E7E6E6} 
\cellcolor[HTML]{E7E6E6} & Faster R-CNN & 0.913 & - & - & - & - & - & - & \cellcolor[HTML]{E7E6E6} \\
\rowcolor[HTML]{E7E6E6} 
\cellcolor[HTML]{E7E6E6} & YOLOv2 & 0.898 & - & - & - & - & - & - & \cellcolor[HTML]{E7E6E6} \\
\rowcolor[HTML]{E7E6E6} 
\multirow{-3}{*}{\cellcolor[HTML]{E7E6E6}\cite{doam}} & SSD300 & 0.915 & - & - & - & - & - & - & \multirow{-3}{*}{\cellcolor[HTML]{E7E6E6}\begin{tabular}[c]{@{}l@{}}- 8150 images\\ - Train to test ratio = 8:2\\ - Objects: knife, gun, shuriken, razor blade\end{tabular}} \\ \hline
\rowcolor[HTML]{FFFFFF} 
\cellcolor[HTML]{FFFFFF} & BoW & - & - & 0.90 & - & - & - & - & \cellcolor[HTML]{FFFFFF} \\
\rowcolor[HTML]{FFFFFF} 
\cellcolor[HTML]{FFFFFF} & Sparse KNN & - & - & 0.95 & - & - & - & - & \cellcolor[HTML]{FFFFFF} \\
\rowcolor[HTML]{FFFFFF} 
\cellcolor[HTML]{FFFFFF} & Sparse KNN* & - & - & 0.89 & - & - & - & - & \cellcolor[HTML]{FFFFFF} \\
\rowcolor[HTML]{FFFFFF} 
\cellcolor[HTML]{FFFFFF} & AISM & - & - & 0.95 & - & - & - & - & \cellcolor[HTML]{FFFFFF} \\
\rowcolor[HTML]{FFFFFF} 
\cellcolor[HTML]{FFFFFF} & XASR+ & - & - & 0.88 & - & - & - & - & \cellcolor[HTML]{FFFFFF} \\
\rowcolor[HTML]{FFFFFF} 
\cellcolor[HTML]{FFFFFF} & GoogleNet & - & - & 0.96 & - & - & - & - & \cellcolor[HTML]{FFFFFF} \\
\rowcolor[HTML]{FFFFFF} 
\cellcolor[HTML]{FFFFFF} & AlexNet & - & - & 0.91 & - & - & - & - & \cellcolor[HTML]{FFFFFF} \\
\rowcolor[HTML]{FFFFFF} 
\cellcolor[HTML]{FFFFFF} & SVM & - & - & 0.86 & - & - & - & - & \cellcolor[HTML]{FFFFFF} \\
\rowcolor[HTML]{FFFFFF} 
\cellcolor[HTML]{FFFFFF} & AdaBoost & - & - & 0.79 & - & - & - & - & \cellcolor[HTML]{FFFFFF} \\
\rowcolor[HTML]{FFFFFF} 
\multirow{-10}{*}{\cellcolor[HTML]{FFFFFF}\cite{S23}} & SRC & - & - & 0.74 & - & - & - & - & \multirow{-10}{*}{\cellcolor[HTML]{FFFFFF}\begin{tabular}[c]{@{}l@{}}- Task: classification \\- Data: B0049-B0051, and B0078-B0082\\ - Classes: gun, shuriken, blade, others\\ \end{tabular}} \\ \hline

\end{tabular}}
\end{table*}

\begin{table*}
\caption{\label{SIXray} \vtop{\hbox{\strut Performance comparison on SIXray dataset}\hbox{\strut $mAP_{cla}$: mean Average Precision (classification), $mAP_{det}$: mean Average Precision (detection)}}}
\centering
\footnotesize
\begin{tabular}{cclcccl}
\hline
\rowcolor[HTML]{AEAAAA} 
\cellcolor[HTML]{AEAAAA} & \cellcolor[HTML]{AEAAAA} & \cellcolor[HTML]{AEAAAA} & \multicolumn{3}{c}{\cellcolor[HTML]{AEAAAA}Performance metric} & \cellcolor[HTML]{AEAAAA} \\
\rowcolor[HTML]{AEAAAA} 
Reference & Method & Subset & \begin{tabular}[c]{@{}l@{}}Localization\\ Accuracy\end{tabular} & $mAP_{cla}$ &$mAP_{det}$ & Experimental protocol \\ \hline
\rowcolor[HTML]{FFFFFF} 
\cellcolor[HTML]{FFFFFF} & ResNet101 & \begin{tabular}[c]{@{}l@{}}SIXray10\\ SIXray100\\ SIXray 1000 \end{tabular} & \begin{tabular}[c]{@{}l@{}}0.501\\ 0.411\\ 0.331\end{tabular} & \begin{tabular}[c]{@{}l@{}}0.774 \\ 0.540\\ 0.360\end{tabular} & \begin{tabular}[c]{@{}l@{}}- \\ -\\ -\end{tabular} & \cellcolor[HTML]{FFFFFF} \\
\rowcolor[HTML]{FFFFFF} 
\cellcolor[HTML]{FFFFFF} & ResNet101+CHR & \begin{tabular}[c]{@{}l@{}}SIXray10\\ SIXray100\\ SIXray 1000 \end{tabular} & \begin{tabular}[c]{@{}l@{}}0.514\\ 0.462\\ 0.398\end{tabular} & \begin{tabular}[c]{@{}l@{}}0.794\\ 0.606\\ 0.381\end{tabular} & \begin{tabular}[c]{@{}l@{}}- \\ -\\ -\end{tabular} & \cellcolor[HTML]{FFFFFF} \\
\rowcolor[HTML]{FFFFFF} 
\cellcolor[HTML]{FFFFFF} & Inspection-v3  & \begin{tabular}[c]{@{}l@{}}SIXray10\\ SIXray100\\ SIXray 1000 \end{tabular}  & \begin{tabular}[c]{@{}l@{}}0.629\\ 0.459\\ 0.303\end{tabular} & \begin{tabular}[c]{@{}l@{}}0.770\\ 0.561\\ 0.387\end{tabular} & \begin{tabular}[c]{@{}l@{}}- \\ -\\ -\end{tabular} & \cellcolor[HTML]{FFFFFF} \\
\rowcolor[HTML]{FFFFFF} 
\cellcolor[HTML]{FFFFFF} & Inspection-v3+CHR & \begin{tabular}[c]{@{}l@{}}SIXray10\\ SIXray100\\ SIXray 1000 \end{tabular}  & \begin{tabular}[c]{@{}l@{}}0.635\\ 0.495\\ 0.315\end{tabular} & \begin{tabular}[c]{@{}l@{}}0.795\\ 0.582\\ 0.469\end{tabular} & \begin{tabular}[c]{@{}l@{}}- \\ -\\ -\end{tabular} &\cellcolor[HTML]{FFFFFF} \\
\rowcolor[HTML]{FFFFFF} 
\cellcolor[HTML]{FFFFFF} & DenseNet & \begin{tabular}[c]{@{}l@{}}SIXray10\\ SIXray100\\ SIXray 1000 \end{tabular}  & \begin{tabular}[c]{@{}l@{}}0.625\\ 0.447\\ 0.346\end{tabular} & \begin{tabular}[c]{@{}l@{}}0.774\\ 0.572\\ 0.393\end{tabular} & \begin{tabular}[c]{@{}l@{}}- \\ -\\ -\end{tabular} & \cellcolor[HTML]{FFFFFF} \\
\rowcolor[HTML]{FFFFFF} 
\multirow{-15}{*}{\cellcolor[HTML]{FFFFFF}\cite{sixray}} & DenseNet+CHR & \begin{tabular}[c]{@{}l@{}}SIXray10\\ SIXray100\\ SIXray 1000 \end{tabular} & \begin{tabular}[c]{@{}l@{}}0.656\\ 0.503\\ 0.439\end{tabular} & \begin{tabular}[c]{@{}l@{}}0.796\\ 0.599\\ 0.484\end{tabular} & \begin{tabular}[c]{@{}l@{}}- \\ -\\ -\end{tabular} & \multirow{-15}{*}{\cellcolor[HTML]{FFFFFF}\begin{tabular}[c]{@{}l@{}}- Tasks: image-level classification (mAP) \\ and localization \\ - 80\% training, 20\% testing\end{tabular}} \\ \hline
\rowcolor[HTML]{E7E6E6} 
\cite{S3} & ResNet50+CST & \begin{tabular}[c]{@{}l@{}}SIXray10\\ SIXray100\\ SIXray 1000 \end{tabular} & \begin{tabular}[c]{@{}l@{}}0.841\\ 0.792\\ 0.752\end{tabular} & \begin{tabular}[c]{@{}l@{}}0.963\\ 0.932\\ 0.890\end{tabular} & \begin{tabular}[c]{@{}l@{}}- \\ -\\ -\end{tabular} & \begin{tabular}[c]{@{}l@{}}- Task: Classification on object proposal level\end{tabular} \\ \hline
\rowcolor[HTML]{FFFFFF} 
\cite{S6} & ResNet50+CST & \begin{tabular}[c]{@{}l@{}}SIXray10\\ SIXray100\\ SIXray 1000 \end{tabular} & \begin{tabular}[c]{@{}l@{}}0.825\\ 0.779\\ 0.743\end{tabular} & \begin{tabular}[c]{@{}l@{}}0.961\\ 0.930\\ 0.889\end{tabular} & \begin{tabular}[c]{@{}l@{}}- \\ -\\ -\end{tabular} & \begin{tabular}[c]{@{}l@{}} - Task: Classification on object proposal level\end{tabular} \\ \hline
\rowcolor[HTML]{E7E6E6} 
\cellcolor[HTML]{E7E6E6} & YOLOv4 & SIXray10 & - & - & 0.701 & \cellcolor[HTML]{E7E6E6} \\
\rowcolor[HTML]{E7E6E6} 
\cellcolor[HTML]{E7E6E6} & DOAM & SIXray10 & - & - & 0.702 & \cellcolor[HTML]{E7E6E6} \\
\rowcolor[HTML]{E7E6E6} 
\cellcolor[HTML]{E7E6E6} & CHR & SIXray10 & - & - & 0.708 & \cellcolor[HTML]{E7E6E6} \\
\rowcolor[HTML]{E7E6E6} 
\cellcolor[HTML]{E7E6E6} & RGBS & SIXray10 & - & -& 0.709 & \cellcolor[HTML]{E7E6E6} \\
\rowcolor[HTML]{E7E6E6} 
\multirow{-5}{*}{\cellcolor[HTML]{E7E6E6}\cite{S015}} & FBS (proposed) & SIXray10 & - & - & 0.712 & \multirow{-5}{*}{\cellcolor[HTML]{E7E6E6}\begin{tabular}[c]{@{}l@{}}- Task: Foreground background separation\\ - 80\% training, 20\% testing\end{tabular}} \\ \hline
\rowcolor[HTML]{FFFFFF} 
\cellcolor[HTML]{FFFFFF} & ResNet-50 & Prohibited items & - & 0.934 & -& \cellcolor[HTML]{FFFFFF} \\
\rowcolor[HTML]{FFFFFF} 
\cellcolor[HTML]{FFFFFF} & ResNet-50 + FPN & Prohibited items & - & 0.939 & - & \cellcolor[HTML]{FFFFFF} \\
\rowcolor[HTML]{FFFFFF} 
\cellcolor[HTML]{FFFFFF} & CHR & Prohibited items & - & 0.940 & - & \cellcolor[HTML]{FFFFFF} \\
\rowcolor[HTML]{FFFFFF} 
\multirow{-4}{*}{\cellcolor[HTML]{FFFFFF}\cite{S7}} & SXMNet (proposed) &  Prohibited items& - & 0.968 &- & \multirow{-4}{*}{\cellcolor[HTML]{FFFFFF}\begin{tabular}[c]{@{}l@{}}- Task: Image-level classification\\ - 7496 training images, 1433 testing images\end{tabular}} \\ \hline
\rowcolor[HTML]{E7E6E6} 
\cellcolor[HTML]{E7E6E6} & CHR \cite{sixray}* & Prohibited items & - & - & 0.794 & \cellcolor[HTML]{E7E6E6} \\
\rowcolor[HTML]{E7E6E6} 
\cellcolor[HTML]{E7E6E6} & Faster R-CNN+FPN & Prohibited items & - & - & 0.795 & \cellcolor[HTML]{E7E6E6} \\
\rowcolor[HTML]{E7E6E6} 
\cellcolor[HTML]{E7E6E6} & Faster R-CNN+IEFPN & Prohibited items & - & - & 0.815 & \cellcolor[HTML]{E7E6E6} \\
\rowcolor[HTML]{E7E6E6} 
\cellcolor[HTML]{E7E6E6} & Cascade R-CNN+FPN & Prohibited items & - & - & 0.813 & \cellcolor[HTML]{E7E6E6} \\
\rowcolor[HTML]{E7E6E6} 
\multirow{-5}{*}{\cellcolor[HTML]{E7E6E6}\cite{wang2021information}} & Cascade R-CNN+IEFPN & Prohibited items & - & - & 0.839 & \multirow{-5}{*}{\cellcolor[HTML]{E7E6E6}\begin{tabular}[c]{@{}l@{}}- Task: Detection\\ - 80\% training, 20\% testing  \\ * result copied from prior work\end{tabular}} \\ \hline
\rowcolor[HTML]{FFFFFF} 
\cellcolor[HTML]{FFFFFF} & YOLOv3 & Prohibited items & - & - & 0.814 & \cellcolor[HTML]{FFFFFF} \\
\rowcolor[HTML]{FFFFFF} 
\cellcolor[HTML]{FFFFFF} & ACMNet & Prohibited items & - & - & 0.843 & \cellcolor[HTML]{FFFFFF} \\
\rowcolor[HTML]{FFFFFF} 
\cellcolor[HTML]{FFFFFF} & YOLOv4 & Prohibited items & - & - & 0.881 & \cellcolor[HTML]{FFFFFF} \\
\rowcolor[HTML]{FFFFFF} 
\multirow{-4}{*}{\cellcolor[HTML]{FFFFFF}\cite{zhou2021x}} & ImprovedYOLOv4 & Prohibited items & - & - & 0.914 & \multirow{-4}{*}{\cellcolor[HTML]{FFFFFF}\begin{tabular}[c]{@{}l@{}}- Task: detection\\ - 7000 training images, 1929 testing images\end{tabular}} \\ \hline
\end{tabular}

\end{table*}

\subsection{GDXray Security}
As shown in Table \ref{GDXray-s}, all available studies on the GDXray Security dataset applied different experiment protocols, such as different subsets of data, different splitting between train and test sets, and different classes included. Therefore, it is not possible to reliably compare the performance of the suggested methods between studies. 

The authors of the GDXray dataset, applied
implicit shape model (ISM), adapted implicit shape model (AISM), \ac{ISM},  \ac{AISM}, \ac{SURF}, and \ac{SIFT}-based non-deep learning methods for threat detection in \cite{S3-15} and AISM led to the highest performance. To evaluate their performance, they computed \ac{ROC} curves on three different \ac{IOU} levels. Then they reported \ac{AUC}, true positive rate (recall) at the false positive rate of 0.05, $R_{0.05}$ as well as true and false positive rates at the best operation point. In Table \ref{GDXray-s}, we report \ac{AUC} and the best operation point results as $R$ and $S_p$ (1-false positive rate), but it should be noted that in the later works also these metrics were computed in a different manner.

In \cite{S3-25}, Faster R-CNN, YOLOv2, and Tiny YOLO were used for object detection, but the performances were evaluated on image level using classification metrics apparently without considering the bounding box overlap in any way. Faster R-CNN achieved the best performance in terms of accuracy, F1 score, recall, and precision. 
The works in \cite{S3, S6} compared the performance of their proposed methods with the methods presented in \cite{S3-25,S3-15}. However, considering different experiment protocols, these comparisons are not reliable as noted also by the authors themselves. 


Although the \ac{MAP} metric was used to assess detection performance in \cite{S14, W08, doam}, it is not possible to reliably compare the performance due to the variations in the experimental protocols. In \cite{S14}, adding PDN branch to Faster R-CNN led to improved performance. In \cite{W08}, a learning-based image synthesis method was proposed to generate more training data. This method was evaluated with four different detection architectures and compared against random training data generation proposed in \cite{saavedra2021generation}. PFPNet was the best-performing architecture, while the proposed image synthesis approach consistently led to better results. In \cite{doam}, different transfer learning techniques were compared, and using SSD300 led to the best performance in comparison to Faster R-CNN and YOLOv2.

Unlike the other studies on the GDXray Security dataset, classification was considered as the \ac{CV} task in \cite{S23}. The performance of ten CV methods was assessed and GoogleNet achieved the best test accuracy. 


\begin{table*}[h!]
\caption{\label{PIDray} \vtop{\hbox{\strut Performance comparison on PIDray dataset}\hbox{\strut \normalfont{$AP$: Average Precision}}}}
\footnotesize
\centering
\begin{tabular}{ccccccccccc}
\hline
\rowcolor[HTML]{A6A6A6} 
\cellcolor[HTML]{A6A6A6}                                     & \cellcolor[HTML]{A6A6A6}                                  & \cellcolor[HTML]{A6A6A6}                                    & \multicolumn{4}{c}{\cellcolor[HTML]{A6A6A6}Detection AP}  & \multicolumn{4}{c}{\cellcolor[HTML]{A6A6A6}Segmentation AP} \\ \cline{4-11} 
\rowcolor[HTML]{A6A6A6} 
\multirow{-2}{*}{\cellcolor[HTML]{A6A6A6}Reference} & \multirow{-2}{*}{\cellcolor[HTML]{A6A6A6}Method} & \multirow{-2}{*}{\cellcolor[HTML]{A6A6A6}Backbone} & Easy & Hard & Hidden & Overall & Easy  & Hard  & Hidden & Overall \\ \hline
                         & FCOS                                                     & ResNet-101-FPN                                              & 61.8          & 51.7          & 37.5            & 50.3             & -              & -              & -               & -                \\
   & RetinaNet                                                & ResNet-101-FPN                                              & 61.8          & 52.2          & 40.6            & 51.5             & -              & -              & -               & -                \\
                              & Faster R-CNN                                               & ResNet-101-FPN                                              & 63.3          & 57.2          & 42.1            & 54.2             & -              & -              & -               & -                \\
    & Libra R-CNN                                               & ResNet-101-FPN                                              & 64.7          & 58.8          & 42.9            & 55.5             & -              & -              & -               & -                \\
                          &  Mask R-CNN                                               & ResNet-101-FPN                                              & 64.7          & 59.0          & 43.8            & 55.8             & 57.6           & 50.2           & 35.2            & 47.7             \\
   & SSD512                                                    & VGG16                                                       & 68.1          & 58.9          & 45.7            & 57.6             & -              & -              & -               & -                \\
                      & Cascade R-CNN                                            & ResNet-101-FPN                                              & 69.3          & 62.8          & 48.0            & 60.0             & -              & -              & -               & -                \\
                    & Cascade Mask R-CNN                                          & ResNet-101-FPN                                              & 70.9          & 64.0          & 48.0            & 61.0             & 59.2           & \textbf{52.2}           & 36.1            & 48.9             \\
\multirow{-9}{*}{\cellcolor[HTML]{FFFFFF}\cite{PIDray}} & SDANet (proposed)                                                    & ResNet-101-FPN                                              & \textbf{71.2}          & \textbf{64.2}          & \textbf{49.5}            & \textbf{61.6}             & \textbf{59.9}           & 52.0           & \textbf{37.4}            & \textbf{49.8}             \\ \hline
\end{tabular}
\end{table*}

\begin{table}
\caption{\label{OPIXray} \vtop{\hbox{\strut Performance comparison on OPIXray dataset} \hbox{\strut \normalfont{ $mAP_{det}$: mean Average Precision (detection)}}\hbox{\strut \normalfont{$mAP_{cla}$: mean Average Precision (classification)}}}}
\centering
\footnotesize
\begin{tabular}{cccc}
\hline
\rowcolor[HTML]{A6A6A6} 
  &   & \multicolumn{2}{l}{Evaluation metric} \\
\rowcolor[HTML]{A6A6A6} 
\multirow{-2}{*}{\cellcolor[HTML]{A6A6A6}Reference}   & \multirow{-2}{*}{\cellcolor[HTML]{A6A6A6}Method}   & $mAP_{det}$ & $mAP_{cla}$  \\ \hline
  & SSD  & 0.709  &- \\
  & SSD+DOAM   & 0.740 &-  \\
  & YOLOv3  & 0.782  &- \\
  & YOLOv3+DOAM  & 0.793 &-   \\
  & FCOS  & 0.820  &- \\
\multirow{-6}{*}{\cite{opixray}} & FCOS+DOAM  & 0.824  &- \\ \hline

\rowcolor[HTML]{E7E6E6}   & YOLOv4 & 0.789 &-  \\
\rowcolor[HTML]{E7E6E6} &  YOLOv4+DOAM   & 0.796 &-  \\
\rowcolor[HTML]{E7E6E6} &  YOLOv4+CHR & 0.786   &- \\
\rowcolor[HTML]{E7E6E6}  &  YOLOv4+RGBS  & 0.790 &- \\
\rowcolor[HTML]{E7E6E6} \multirow{-5}{*}{\cite{S015}}  &  YOLOv4+FBS (proposed)  & 0.818  &- \\ \hline
  & SSD+LIM   & 0.746 &-  \\
  & FCOS+LIM   & 0.831  &- \\
  & YOLOv5   & 0.878  &- \\ 
  & YOLOv5+DOAM   & 0.888 &-  \\  
\multirow{-5}{*}{\cite{HiXray}}  & YOLOv5+LIM   & 0.906  &- \\ \hline
\rowcolor[HTML]{E7E6E6}  & ATSS   & 0.866 &-  \\
\rowcolor[HTML]{E7E6E6}  & ATSS+DOAM   & 0.856   &- \\
\rowcolor[HTML]{E7E6E6} & ATSS+LAreg (proposed)   & 0.874  &- \\
\rowcolor[HTML]{E7E6E6}
\multirow{-2}{*}{\cite{CLCXray}}   & ATSS+LAcls (proposed)  & 0.883 &-  \\ \hline
  & Faster R-CNN+FPN & 0.801  &- \\
  & Faster R-CNN+IEFPN   & 0.817 &-  \\
  & Cascade R-CNN+FPN  & 0.780 &- \\ 
\multirow{-4}{*}{\cite{wang2021information}} & Cascade R-CNN+IEFPN  & 0.799   \\ \hline
\rowcolor[HTML]{E7E6E6} 
\cellcolor[HTML]{E7E6E6}  & ResNet-50 &-  & 0.864   \\
\rowcolor[HTML]{E7E6E6} 
\cellcolor[HTML]{E7E6E6}  & ResNet-50+FPN &- & 0.866   \\
\rowcolor[HTML]{E7E6E6} 
\cellcolor[HTML]{E7E6E6}  & CHR &- & 0.877   \\
\rowcolor[HTML]{E7E6E6} 
\multirow{-4}{*}{\cellcolor[HTML]{E7E6E6}\cite{S7}}  & SXMNet (proposed) &-  & 0.908   \\ \hline
\end{tabular}
\end{table}

\subsection{SIXray}
As explained in \ref{ssec:sixray}, the original dataset paper \cite{sixray} defines three different subsets of data (SIXray10, SIXray100, and SIXray1000) as well as two tasks: image-level classification evaluated by \ac{MAP} and object localization evaluated by localization accuracy. The original dataset paper also provides baseline results for three different network architectures with/without class-balanced hierarchical refinement (CHR) on these tasks on each subset as reported in Table \ref{SIXray}.

In \cite{S3, S6}, the performance is directly compared against results copied from \cite{sixray}, but due to a different experimental protocol, this is questionable. In \cite{S3, S6}, the classification is carried out on object proposal level, not on image-level as in \cite{sixray}. Furthermore, while not clearly described, it appears that only the detected object proposals are considered, i.e., completely undetected objects will not harm the classification performance. The computation of localization accuracy is not described.  

In \cite{S015}, the focus is on foreground-background separation. The proposed method FBS is compared against four detection methods, namely, YOLOv4, De-Occlusion Attention Module (DOAM), CHR, and RGBS on SIXray10 data using \ac{MAP} (detection) at 0.5 \ac{IOU} as the evaluation metric.

The remaining studies \cite{S7,wang2021information,zhou2021x} on SIXray dataset did not use the subsets defined in \cite{sixray}, but instead, they picked only the images containing prohibitive items (8929 images). Image-level multi-label classification was tackled in \cite{S7}, whereas  \cite{wang2021information,zhou2021x} focused on detection. In \cite{wang2021information}, the results are directly compared with image-level classification results picked from \cite{sixray} despite the different subset of the dataset used and the different \ac{CV} task evaluated.  Furthermore, different splitting into training and test sets makes the results from the detection papers \cite{wang2021information,zhou2021x} incomparable. Furthermore, the papers do not report the \ac{IOU} threshold used for \ac{MAP} evaluation, which makes also future comparisons with the reported results unreliable.

\subsection{PIDray}
At the time of writing this paper, the main paper that introduced PIDray dataset \cite{PIDray} was the only one reporting results on this dataset. Two tasks, i.e., detection and segmentation were considered and several methods were evaluated for both tasks using \ac{AP} obtained by averaging over multiple \ac{IOU} levels and all categories. We report these results in Table \ref{PIDray}.

The methods evaluated in \cite{PIDray} included the proposed Selective Dense attention Network (SDANet), which is an architecture based on Cascade Mask-RCNN \cite{cai2019cascade} that has a ResNet-101 network as its backbone. It can be seen that the SDANet achieved the best performance compared to others with overall Detection AP and Segmentation AP of 61.6 and 49.8, respectively.

\subsection{OPIXray}
The original OPIXray dataset paper \cite{opixray} provides several results for object detection task evaluated using \ac{MAP} at 0.5 \ac{IOU} threshold focusing on the performance of the proposed De-occlusion Attention Module (DOAM). In Table \ref{OPIXray}, we report the results for three architectures (SSD, YOLOv3, and fully-convolution one-stage object detector (FCOS) \cite{tian2019fcos}) with and without DOAM. FCOS+DOAM obtained the best performance. 

Compared to other datasets, there appears to be more consistency in the way the OPIXray dataset has been used in the evaluations, and therefore, comparison of results across papers is possible.
In \cite{S015}, YOLOv4 was evaluated by itself and with different additional modules, namely DOAM, CHR, RGBS, and the proposed FBS. The proposed FBS approach achieved the best performance in terms of \ac{MAP} (shown in Table \ref{OPIXray}), $\eta$, and Recall, while DOAM achieved the best performance in terms of precision and F1 metrics. In \cite{HiXray}, the authors of the OPIXray dataset paper proposed Lateral Inhibition Module (LIM) and provided some additional results that are reliably comparable with the original OPIXray results. The best performance was achieved by YOLOv5 combined with LIM. Adaptive Training Sample
Selection (ATSS) model \cite{zhang2020bridging} was evaluated in \cite{CLCXray} by itself, with DOAM, and with two proposed Label-aware Mechanisms. Lable-aware classification (LAcls) achieved the best performance. In \cite{wang2021information}, Cascade R-CNN + Information-exchange Enhanced Feature Pyramid Network (IEFPN) led to the best accuracy in comparison to other implemented methods.

The dataset has been also used for image-level multi-label classification in \cite{S7}. The performance of the proposed SXMNet was compared with ResNet-50, ResNet-50 + FPN, and CHR. As the results show, the suggested method achieved the best performance. 

\begin{table}
\caption{\label{HiXray} \vtop{\hbox{\strut Performance comparison on HiXray dataset}\hbox{\strut \normalfont{$mAP$: mean Average Precision}}}}
\centering
\footnotesize
\begin{tabular}{ccc}
\hline
\rowcolor[HTML]{A6A6A6} 
\cellcolor[HTML]{A6A6A6} & \cellcolor[HTML]{A6A6A6} & \multicolumn{1}{l}{\cellcolor[HTML]{A6A6A6}Evaluation metrics} \\
\rowcolor[HTML]{A6A6A6} 
\multirow{-2}{*}{\cellcolor[HTML]{A6A6A6}Reference} & \multirow{-2}{*}{\cellcolor[HTML]{A6A6A6}Method} & $mAP_{bbox}$ \\ \hline
 & SSD & 0.714 \\
 & SSD +DOAM & 0.721 \\
 & SSD+LIM & 0.731 \\
  & FCOS & 0.757 \\
   & FCOS+DOAM & 0.762 \\
 & FCOS+LIM & 0.773 \\
   & YOLOv5 & 0.817 \\
   & YOLOv5+DOAM & 0.822 \\
\multirow{-9}{*}{\cite{HiXray}} & \textbf{YOLOv5+LIM} & {\color[HTML]{000000} \textbf{0.832}} \\ \hline
\end{tabular}
\end{table}

\begin{table*}
\caption{\label{CLCXray} \vtop{\hbox{\strut Performance comparison on CLCXray dataset}\hbox{\strut \normalfont{For evaluation metric definitions, see Section \ref{ssec:clcxray}}}}}
\centering
\footnotesize
\begin{tabular}{cccccccc}
\hline
\rowcolor[HTML]{A6A6A6} 
\cellcolor[HTML]{A6A6A6}                                                & \cellcolor[HTML]{A6A6A6}                           & \multicolumn{6}{c}{\cellcolor[HTML]{A6A6A6}Evaluation metrics}                                            \\
\rowcolor[HTML]{A6A6A6} 
\multirow{-2}{*}{\cellcolor[HTML]{A6A6A6}Reference}                     & \multirow{-2}{*}{\cellcolor[HTML]{A6A6A6}Method}   & $mAP$  & $mAP_50$  & $mAP_75$  & $mAP_{s}$             & $mAP_{m}$       & $mAP_{l}$                  \\ \hline
    & SSD  & $51.1\pm0.3$    & $66.4\pm0.3$ &$59.8\pm0.4$                                & $0.7\pm1.0$              & $22.0\pm0.5$              & $57.5\pm0.4$                            \\
    & YOLOv3 & $53.0\pm0.1$      & $67.2\pm0.3$ & $63.0\pm0.2$                                & $0.0\pm0.0$              & $25.9\pm2.1$              & $58.6\pm0.6$                            \\
& FCOS    & $56.3\pm0.0$    & $70.7\pm0.2$  & $66.6\pm0.4$ &   \textbf{36.3$\pm$6.0}              & $27.3\pm2.3$              & $62.1\pm0.3$                            \\
& NAS-FCOS    & $57.3\pm0.1$    & \textbf{72.3$\pm$0.4}  & $67.7\pm0.5$ &   30.3$\pm$2.8              & $28.8\pm0.8$              & $63.3\pm0.2$                            \\
& PAA    & $58.3\pm0.1$    & $71.6\pm0.4$  & \textbf{68.5.$\pm$0.3} &   19.8$\pm$3.4              & $29.4\pm1.2$              & $63.9\pm0.1$                            \\
& ATSS     &$58.0\pm0.2$    & $70.8\pm0.0$   & $67.2\pm0.2$                              & $17.9\pm4.1$              & $31.0\pm0.4$              & $63.3\pm0.0$                            \\
& ATSS+LAreg (proposed) & $58.5\pm0.1$ & $70.9\pm0.1$    & $67.7\pm0.5$                           & $12.6\pm4.7$              & $30.5\pm0.8$                            & $63.8\pm0.2$              \\
\multirow{-8}{*}{\cite{tao2021over}}                           & ATSS+LAcls (proposed)             & \textbf{59.3$\pm$0.2}            & 71.8$\pm$0.2 & $68.2\pm0.1$ &  23.0$\pm$10.5                            & \textbf{32.4$\pm$0.5}              & \textbf{64.5$\pm$0.1 }             \\ \hline
\end{tabular}
\end{table*}

\subsection{HiXray}

At the time of preparing this article, the HIXray dataset paper \cite{HiXray} was the only one providing results on the dataset. Three object detection architectures (SSD, FCOS, YOLOv5) were evaluated as such, with DOAM and with the proposed LIM. The performance was evaluated using \ac{MAP} at 0.5 \ac{IOU}. 
As shown in Table \ref{HiXray}, the combination of YOLOv5 and LIM led to the highest performance.

\subsection{CLCXray}

At the time of writing this article, only the CLCXray dataset paper \cite{CLCXray} provides results on the CLCXray dataset. Several approaches for object detection are evaluated using the COCO evaluation metrics as described in Section \ref{ssec:clcxray}. We report results for the best-performing approaches along with some results for well-known detection architectures in Table \ref{CLCXray}. Based on the main metric $mAP$ the proposed method using LAcls reached to the best performance.

\section{Conclusion}
\label{sec:conclusion}

As a non-destructive technology, X-ray imaging is finding use in different industrial and security applications to assess the inner structure or contents by measuring mass distributions (absorption rate). Automatic assessment of X-ray images, in terms of detection, classification, and segmentation, can be achieved by applying CV-based methods. In this paper, a review of CV studies on X-ray data applications in industrial production and security areas was presented.

While a large number of recent studies have focused on this topic and many advances have been made as evident from our review, we observed a lot of room for improvement for the field as a whole in experimental evaluation. To advance as a field, the proposed approaches and obtained results need to be comparable across studies. At the moment, a large part of the studies use only private data. Furthermore, even the studies using public datasets use varying experimental setups that often make the comparisons incomparable. 

We recommend to all the works comparing their results with prior works to first carefully check the computer vision task and note that the same evaluation metrics can be used in classification, detection, and segmentation, while the results naturally are not comparable. Furthermore, we recommend carefully following the experimental protocols including data subsets, splitting into training and testing sets, and evaluation metrics and note that any variations make the results incomparable. It should be also noted that common metrics, such as mean average precision have multiple implementations, and therefore, every work should carefully report also the details of their selected metric. 

New larger datasets for different applications are still needed. For newly published datasets, it is naturally important to document all the details of the experimental protocol. To avoid variations in later works using the same dataset, we also recommend publishing the implementations for running evaluations on the dataset.

\bibliographystyle{ieeetr}
\bibliography{access}


\EOD

\end{document}